\documentclass[10pt,twocolumn,letterpaper]{article}

\usepackage{cvpr}
\usepackage{times}
\usepackage{epsfig}
\usepackage{enumitem}
\usepackage{graphicx}
\usepackage{amsmath}
\usepackage{amssymb}
\usepackage{bm}
\usepackage{wrapfig}
\usepackage{caption}
\usepackage{subcaption}
\usepackage{enumitem}
\usepackage{color}
\usepackage{xspace}

\newcommand*{\permcomb}[4][0mu]{{{}^{#3}\mkern#1#2_{#4}}}
\newcommand*{\comb}[1][-1mu]{\permcomb[#1]{C}}
\newcommand{\mathcider}{\text{CIDEr}\xspace}
\newcommand{\mathciderD}{\text{CIDEr-D}\xspace}

\newcommand{\bleu}{BLEU\xspace}
\newcommand{\bleux}{BLEU}
\newcommand{\rouge}{ROUGE\xspace}
\newcommand{\rougex}{ROUGE}
\newcommand{\cider}{CIDEr\xspace}
\newcommand{\ciderx}{CIDEr}
\newcommand{\meteor}{METEOR\xspace}
\newcommand{\rs}{s\xspace}
\newcommand{\RS}{S\xspace}
\newcommand{\ngram}{$n$-gram\xspace}
\newcommand{\ngrams}{$n$-grams\xspace}
\newcommand{\rougen}{ROUGE$_N$\xspace}
\newcommand{\rougel}{ROUGE$_L$\xspace}
\newcommand{\rouges}{ROUGE$_S$\xspace}
\newcommand{\cidern}{CIDEr$_n$\xspace}
\newcommand{\pascal}{PASCAL-50S\xspace}
\newcommand{\abstractscenes}{ABSTRACT-50S\xspace}
\newcommand{\coco}{MS COCO\xspace}

\newenvironment{packed_itemize}{
\begin{list}{\labelitemi}{\leftmargin=2em}
\vspace{-6pt}
  \setlength{\itemsep}{0pt}
  \setlength{\parskip}{0pt}
  \setlength{\parsep}{0pt}
}{\end{list}}
\usepackage[pagebackref=true,breaklinks=true,letterpaper=true,colorlinks,bookmarks=false]{hyperref}
\cvprfinalcopy % *** Uncomment this line for the final submission
\usepackage[author={DP}, open=true]{pdfcomment}

\def\cvprPaperID{39} % *** Enter the CVPR Paper ID here

\ifcvprfinal\fi
\begin{document}

%%%%%%%%% TITLE
\title{CIDEr: Consensus-based Image Description Evaluation}

\author{Ramakrishna Vedantam\\
Virginia Tech\\
{\tt\small vrama91@vt.edu}
\and
C. Lawrence Zitnick\\
Microsoft Research\\
{\tt\small larryz@microsoft.com}
\and
Devi Parikh\\
Virgnia Tech\\
{\tt\small parikh@vt.edu}
}

\maketitle
%\thispagestyle{empty}

%%%%%%%%% ABSTRACT
\begin{abstract}
Automatically describing an image with a sentence is a long-standing challenge in computer vision and natural language processing. Due to recent progress in object detection, attribute classification, action recognition, \etc., there is renewed interest in this area. However, evaluating the quality of descriptions has proven to be challenging. We propose a novel paradigm for evaluating image descriptions that uses \emph{human consensus}. This paradigm consists of three main parts: a new triplet-based method of collecting human annotations to measure consensus, a new automated metric (\cider) that captures consensus, and two new datasets: \pascal and \abstractscenes that contain 50 sentences describing each image. Our simple metric captures human judgment of consensus better than existing metrics across sentences generated by various sources. We also evaluate five state-of-the-art image description approaches using this new protocol and provide a benchmark for future comparisons. A version of \cider named \ciderx-D is available as a part of \coco evaluation server to enable systematic evaluation and benchmarking.
\end{abstract} 
%%%%%%%%% BOD\begin{flushright}

\section{Introduction}
Recent advances in object recognition \cite{DPM}, attribute classification \cite{Lampert}, action classification \cite{MajiActionCVPR11,Dokania_Ranking_ECCV14} and crowdsourcing \cite{4562953} have increased the interest in solving higher level scene understanding problems. One such problem is generating human-like descriptions of an image. In spite of the growing interest in this area, the evaluation of novel sentences generated by automatic approaches remains challenging. Evaluation is critical for measuring progress and spurring improvements in the state of the art. This has already been shown in various problems in computer vision, such as detection~\cite{pascal-voc-2010,imagenet_cvpr09}, segmentation~\cite{pascal-voc-2010,MartinFTM01}, and stereo~\cite{Scharstein:2002:TED:598429.598475}.

Existing evaluation metrics for image description attempt to measure several desirable properties. These include grammaticality, saliency (covering main aspects), correctness/truthfulness, \etc. Using human studies, these properties may be measured, \eg on separate {\it one} to {\it five}~\cite{midge,rohrbach13iccv, YangTDA11,ElliottK13} or \emph{pairwise} scales~\cite{Yatskar}. Unfortunately, combining these various results into one measure of sentence quality is difficult. Alternatively, other works \cite{babytalk,journals/jair/HodoshYH13} ask subjects to judge the overall quality of a sentence. 

% The inherent subjectivity of this task leads to uncertainty in how the subjects weight the various sentence properties when making their final judgments. Different subjects may also use different criteria. These factors make it difficult to evaluate progress reliably.

\begin{figure*}[t]
\includegraphics[width=\textwidth]{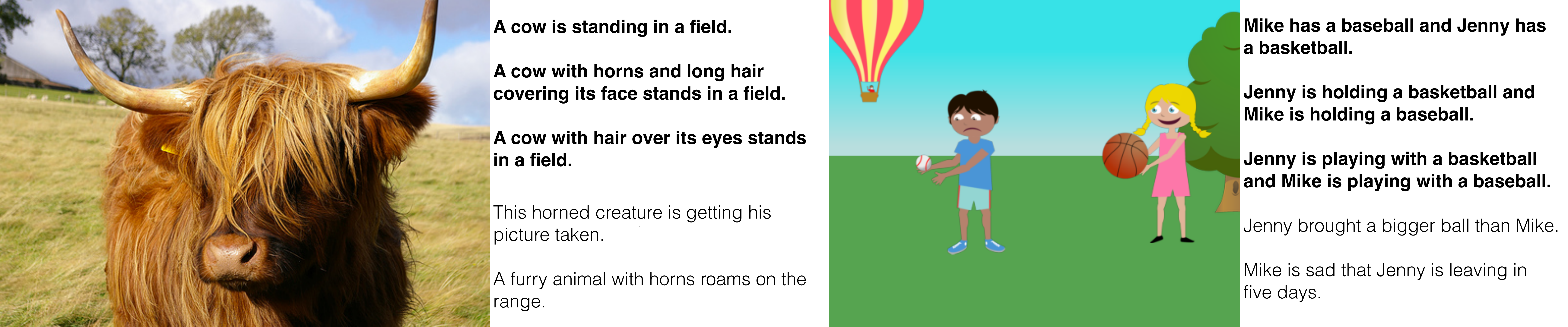}
\caption{Images from our \pascal (left) and \abstractscenes (right) datasets with a subset of corresponding (human) sentences. Sentences shown in \textbf{bold} are representative of the consensus descriptions for these images. We propose to capture such descriptions with our evaluation protocol.}
\vspace{-10pt}
\label{Fig:1}
\end{figure*}

An important yet non-obvious property exists when image descriptions are judged by humans: What humans like often does not correspond to what is human-like.\footnote{This is a subtle but important distinction. We show qualitative examples of this in the appendix. That is, the sentence that is most similar to a typical human generated description is often not judged to be the ``best'' description. In this paper, we propose to directly measure the ``human-likeness'' of automatically generated sentences.}
%Common to all of these approaches is the desire to measure the ``human-likeness'' of a sentence~\cite{Li:2011:CSI:2018936.2018962,elliott-keller:2014:P14-2}. 
%\pdfcomment{I am generally more ok with somewhat bold statements than Larry usually is -- most of the time when writing intros, Larry would tone some statements down, and I'd bolden them up a bit till we converge or agree to disagree :-)}
%\larry{Damn, that was a bold sentence.} \rama{Maybe pick off from the previous paragraph and say this instead: To evaluate progress on generating more human-like descriptions, we propose a more grounded approach.}
We introduce a novel consensus-based evaluation protocol, which measures the similarity of a sentence to the majority, or {\em consensus} of how most people describe the image (Fig.~\ref{Fig:1}).
%\rama{To me it still feels like a leap from human-like to consensus. Maybe we can describe the interface first, which clearly captures human-like, and then show how over more references it captures consensus?}
%\pdfcomment{Doesn't feel like a leap to me. Why does it feel like a leap to you?}
One realization of this evaluation protocol uses human subjects to judge sentence similarity between a candidate sentence and human-provided ground truth sentences. The question ``Which of two sentences is more similar to this other sentence?'' is posed to the subjects. The resulting quality score is based on how often a sentence is labeled as being \emph{more} similar to a human-generated sentence. The relative nature of the question helps make the task objective. We encourage the reader to review how a similar protocol has been used in~\cite{Tamuz11adaptivelylearning} to capture human perception of image similarity. These annotation protocols for similarity may be understood as instantiations of 2AFC (two alternative forced choice) ~\cite{BogaczBrownMoehlisEtAl06}, a popular modality in psychophysics. %\rama{Justifying triplets with both citations makes it feel weak. So, I changed the tone of the first citation.}

%\rama{We encourage the reader to review, how similar ideas have been used for measuring human perception of image similarity~\cite{add}.}\larry{Why isn't this subjective as well?}

Since human studies are expensive, hard to reproduce, and slow to evaluate, automatic evaluation measures are commonly desired. To be useful in practice, automated metrics should agree well with human judgment. Some popular metrics used for image description evaluation are \bleu~\cite{papineni} (precision-based) from the machine translation community and \rouge~\cite{rouge} (recall-based) from the summarization community. Unfortunately, these metrics have been shown to correlate weakly with human judgment~\cite{babytalk, ElliottK13, Callison-burch06re-evaluatingthe, journals/jair/HodoshYH13}. For the
%subjective
task of judging the overall quality of a description, the \meteor~\cite{ElliottK13} metric has shown better correlation with human subjects. Other metrics rely on the ranking of captions \cite{journals/jair/HodoshYH13} and cannot evaluate novel image descriptions.

We propose a new automatic {\em consensus} metric of image description quality -- \cider (Consensus-based Image Description Evaluation). Our metric measures the similarity of a generated sentence against a set of ground truth sentences written by humans. Our metric shows high agreement with consensus as assessed by humans. Using sentence similarity, the notions of grammaticality, saliency, importance and accuracy (precision and recall) are inherently captured by our metric.
% \larry{Should we mention this earlier?}

Existing datasets popularly used to evaluate image description approaches have a maximum of only five descriptions per image \cite{Rashtchian:2010:CIA:1866696.1866717,journals/jair/HodoshYH13,Ordonez:2011:im2text}. However, we find that five sentences are not sufficient for measuring how a ``majority" of humans would describe an image. Thus, to accurately measure consensus, we collect two new evaluation datasets containing 50 descriptions per image -- \pascal and \abstractscenes. The \pascal dataset is based on the popular UIUC Pascal Sentence Dataset, which has 5 descriptions per image. This dataset has been used for both training and testing in numerous works~\cite{midge, babytalk, Story, rohrbach13iccv}. The \abstractscenes dataset is based on the dataset of Zitnick and Parikh~\cite{ZitnickCVPR2013}. While previous methods have only evaluated using 5 sentences, we explore the use of 1 to {\raise.17ex\hbox{$\scriptstyle\sim$}}50 reference sentences. Interestingly, we find that most metrics improve in performance with more sentences.\footnote{Except \bleu computed on unigrams} Inspired by this finding, the \coco testing dataset now contains 5K images with 40 reference sentences to boost the accuracy of automatic measures~\cite{2015arXiv150400325C}.

{\bf Contributions:} In this work, we propose a consensus-based evaluation protocol for image descriptions. We introduce a new annotation modality for human judgment, a new automated metric, and two new datasets. We compare the performance of five state-of-the-art machine generation approaches~\cite{midge,babytalk,Story, rohrbach13iccv}. %\rama{I have removed the images with consensus descriptions part from the paper, for lack of space. Can we sneak it in somewhere or put it in the supplementary?}
%Further, we notice that some images tend to have more consensus sentences than others.We study such images on \abstractscenes, \pascal and the newly released Common Objects in COntext (COCO) dataset~\cite{add}.
Our code and datasets are available on the author's webpages. Finally, to facilitate the adoption of this protocol, we have made \cider available as a metric on the newly released \coco caption evaluation server~\cite{2015arXiv150400325C}. 
\section{Related Work}
{\bf Vision and Language}:
Numerous papers have studied the relationship between language constructs and image content. Berg~\etal~\cite{conf/cvpr/BergBDDGHMMSSY12} characterize the relative importance of objects (nouns). Zitnick and Parikh~\cite{ZitnickCVPR2013} study relationships between visual and textual features by creating a synthetic Abstract Scenes Dataset. Other works have modeled prepositional relationships~\cite{conf/eccv/GuptaD08}, attributes (adjectives)~\cite{Lampert,citeulike:10025772}, and visual phrases (\ie visual elements that co-occur)~\cite{VisualPhrases}. Recent works have utilized techniques in deep learning to learn joint embeddings of text and image fragments~\cite{DBLP:journals/corr/KarpathyJF14}.
\begin{figure*}
\centering
\begin{subfigure}{0.3\textwidth}
\includegraphics[width=\textwidth, page=1]{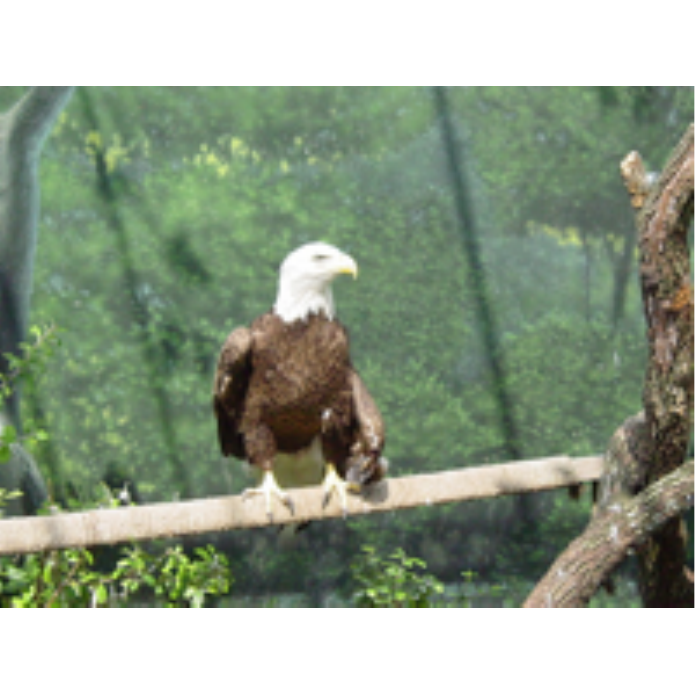}
\caption{}
\end{subfigure}
\begin{subfigure}{0.3\textwidth}
\includegraphics[width=\textwidth, page=2]{figures/fig_2}
\caption{}
\end{subfigure}
\begin{subfigure}{0.3\textwidth}
\includegraphics[width=\textwidth, page=3]{figures/fig_2}
\caption{}
\end{subfigure}
\vspace{-10pt}
\caption{Illustration of our triplet annotation modality. Given an image (a), with reference sentences (b) and a pair of candidate sentences (c, top), we match them with a reference sentence one by one to form triplets (c, bottom). Subjects are shown these 50 triplets on Amazon Mechanical Turk and asked to pick which sentence (B or C) is more similar to sentence A. }
\vspace{-10pt}
\label{fig:triplet_interface}
\end{figure*}

{\bf Image Description Generation}:
Various methods have been explored for generating full descriptions for images. Broadly, the techniques are either retrieval-~\cite{Story,Ordonez:2011:im2text,journals/jair/HodoshYH13} or generation-based~\cite{midge,babytalk,Yatskar,rohrbach13iccv}. While some retrieval-based approaches use global retrieval~\cite{Story}, others retrieve text phrases and stitch them together in an approach inspired by extractive summarization~\cite{Ordonez:2011:im2text}. Recently, generative approaches based on combination of Convolutional and Recurrent Neural Networks~\cite{DBLP:journals/corr/KarpathyF14,DBLP:journals/corr/ChenZ14a,DBLP:journals/corr/DonahueHGRVSD14,DBLP:journals/corr/VinyalsTBE14,DBLP:journals/corr/MaoXYWY14,DBLP:journals/corr/KirosSZ14} have created a lot of excitement. Other generative approaches have explored creating sentences by inference over image detections and text-based priors~\cite{babytalk} or exploiting word co-occurrences using syntactic trees~\cite{midge}. Rohrbach \etal~\cite{rohrbach13iccv} propose a machine translation approach that goes from an intermediate semantic representation to sentences. %Yang \etal propose a slot filling approach~\cite{YangTDA11} based on methods used for text summarization.
Some other approaches include~\cite{Jawahar,Li:2011:CSI:2018936.2018962,YangTDA11,Yatskar}. Most of the approaches use the UIUC Pascal Sentence ~\cite{Story,babytalk,midge,rohrbach13iccv,Jawahar} and the \coco datasets~\cite{DBLP:journals/corr/KarpathyF14,DBLP:journals/corr/ChenZ14a,DBLP:journals/corr/DonahueHGRVSD14,DBLP:journals/corr/VinyalsTBE14,DBLP:journals/corr/MaoXYWY14,DBLP:journals/corr/KirosSZ14} for evaluation.
%Recent work has also looked at generating descriptions of images with dense human visual annotation~\cite{Yatskar}.
In this work we focus on the problem of evaluating image captioning approaches.

{\bf Automated Evaluation}:
Automated evaluation metrics have been used in many domains within Artificial Intelligence (AI), such as statistical machine translation and text summarization. Some of the popular metrics in machine translation include those based on precision, such as \bleu~\cite{papineni} and those based on precision as well as recall, such as \meteor~\cite{meteor}. While BLEU (BiLingual Evaluation Understudy) has been the most popular metric, its effectiveness has been repeatedly questioned~\cite{babytalk, ElliottK13, Callison-burch06re-evaluatingthe, journals/jair/HodoshYH13}. A popular metric in the summarization community is ROUGE~\cite{rouge} (Recall Oriented Understudy of Gisting Evaluation). This metric is primarily recall-based and thus has a tendency to reward long sentences with high recall. These metrics have been shown to have weak to moderate correlation with human judgment~\cite{ElliottK13}. Recently, \meteor has been used for image description evaluation with more promising results~\cite{elliott-keller:2014:P14-2}. Another metric proposed by Hodosh \etal~\cite{journals/jair/HodoshYH13} can only evaluate ranking-based approaches, it cannot evaluate novel sentences. We propose a consensus-based metric that rewards a sentence for being similar to the majority of human written descriptions. Interestingly, similar ideas have been used previously to evaluate text summarization~\cite{conf/naacl/NenkovaP04}.

{\bf Datasets}:
Numerous datasets have been proposed for studying the problem of generating image descriptions. The most popular dataset is the UIUC Pascal Sentence Dataset~\cite{Rashtchian:2010:CIA:1866696.1866717}. This dataset contains 5 human written descriptions for 1,000 images. This dataset has been used by a number of approaches for training and testing. The SBU captioned photo dataset \cite{Ordonez:2011:im2text} contains one description per image for a million images, mined from the web. These are commonly used for training image description approaches. Approaches are then tested on a query set of 500 images with one sentence each. The Abstract Scenes dataset~\cite{ZitnickCVPR2013} contains cartoon-like images with two descriptions. The recently released \coco dataset~\cite{LinECCV14coco} contains five sentences for a collection of over 100K images. This dataset is gaining traction with recent image description approaches~\cite{DBLP:journals/corr/KarpathyF14,DBLP:journals/corr/ChenZ14a,DBLP:journals/corr/DonahueHGRVSD14,DBLP:journals/corr/VinyalsTBE14,DBLP:journals/corr/MaoXYWY14,DBLP:journals/corr/KirosSZ14}. Other datasets of images and associated descriptions include ImageClef~\cite{Mller:2010:IEE:1869912} and Flickr8K~\cite{journals/jair/HodoshYH13}. In this work, we introduce two new datasets. First is the \pascal dataset where we collected 50 sentences per image for the 1,000 images from UIUC Pascal Sentence dataset. The second is the \abstractscenes dataset where we collected 50 sentences for a subset of 500 images from the Abstract Scenes dataset. We demonstrate that more sentences per image are essential for reliable automatic evaluation.

The rest of this paper is organized as follows. We first give details of our triplet human annotation modality (Sec.~\ref{sec:human_studies}). Then we provide the details of our consensus-based automated metric, CIDEr (Sec.~\ref{sec:metric}). In Sec.~\ref{sec:dataset} we provide the details of our two new image-sentence datasets, \pascal and \abstractscenes. Our contributions of triplet annotation, metric and dataset make consensus-based image description evaluation feasible. Our results (Sec.~\ref{sec:results}) demonstrate that our automated metric and our proposed datasets capture consensus better than existing choices.

All our human studies are performed on the Amazon Mechanical Turk (AMT). Subjects are restricted to the United States, and other qualification criteria are imposed based on worker history.\footnote{Approval rate greater than 95\%, minimum 500 HITs approved}
\section{Consensus Interface}
\label{sec:human_studies}
Given an image and a collection of human generated \emph{reference} sentences describing it, the goal of our consensus-based protocol is to measure the similarity of a \emph{candidate} sentence to a majority of how most people describe the image (\ie the \emph{reference} sentences). In this section, we describe our human study protocol for generating ground truth consensus scores. In Sec.~\ref{sec:results}, these ground truth scores are used to evaluate several automatic metrics including our proposed \cider metric.

An illustration of our human study interface is shown in Fig.~\ref{fig:triplet_interface}. Subjects are shown three sentences: A, B and C. They are asked to pick which of two sentences (B or C) is most similar to sentence A. Sentences B and C are two candidate sentences, while sentence A is a reference sentence. For each choice of B and C, we form triplets using all the reference sentences for an image. We provide no explicit concept of ``similarity". Interestingly, even though we do not say that the sentences are image descriptions, some workers commented that they were imagining the scene to make the choice. The relative nature of the task -- ``Which of the two sentences, B or C, is more similar to A?'' -- helps make the assessment more objective. That is, it is easier to judge if one sentence is more similar than another to a sentence, than to provide an absolute rating from 1 to 5 of the similarity between two sentences \cite{BogaczBrownMoehlisEtAl06}.

We collect three human judgments for each triplet. For every triplet, we take the majority vote of the three judgments. For each pair of candidate sentences (B, C), we assign B the winner if it is chosen as more similar by a majority of triplets, and similarly for C. These pairwise relative rankings are used to evaluate the performance of the automated metrics. % described next.
That is, when automatic metrics give both sentences B and C a score, we check whether B received a higher score or C. Accuracy is computed as the proportion of candidate pairs on which humans and the automatic metric agree on which of the two sentences is the winner.

\section{\cider Metric}
\label{sec:metric}

Our goal is to automatically evaluate for image $I_i$ how well a candidate sentence $c_i$ matches the consensus of a set of image descriptions $\RS_{i} = \{\rs_{i1},\ldots,\rs_{im}\}$. All words in the sentences (both candidate and references) are first mapped to their stem or root forms.  That is, ``fishes'', ``fishing'' and ``fished'' all get reduced to ``fish.'' We represent each sentence using the set of \ngrams present in it. An \ngram~$\omega_k$ is a set of one or more ordered words. In this paper we use \ngrams~containing one to four words.

Intuitively, a measure of consensus would encode how often \ngrams~in the candidate sentence are present in the reference sentences. Similarly, \ngrams~not present in the reference sentences should not be in the candidate sentence. Finally, \ngrams~that commonly occur across all images in the dataset should be given lower weight, since they are likely to be less informative. To encode this intuition, we perform a Term Frequency Inverse Document Frequency (TF-IDF) weighting for each \ngram~\cite{Robertson04understandinginverse}. The number of times an \ngram~$\omega_k$ occurs in a reference sentence $\rs_{ij}$ is denoted by $h_k(\rs_{ij})$ or $h_k(c_i)$ for the candidate sentence $c_i$. We compute the TF-IDF weighting $g_k(\rs_{ij})$ for each \ngram~$\omega_k$ using:
\vspace{-5pt}
\begin{multline}
g_k(\rs_{ij}) = \\ \frac{h_k(\rs_{ij})}{\sum_{\omega_l \in \Omega} h_l(\rs_{ij})} \log\left(\frac{|I|}{\sum_{I_p \in I} \min(1, \sum_q h_k(s_{pq}))}\right),
\end{multline}

where $\Omega$ is the vocabulary of all \ngrams and $I$ is the set of all images in the dataset. The first term measures the TF of each \ngram~$\omega_k$, and the second term measures the rarity of $\omega_k$ using its IDF.  Intuitively, TF places higher weight on \ngrams~that frequently occur in the reference sentence describing an image, while IDF reduces the weight of \ngrams~that commonly occur across all images in the dataset. That is, the IDF provides a measure of word saliency by discounting popular words that are likely to be less visually informative. The IDF is computed using the logarithm of the number of images in the dataset $|I|$ divided by the number of images for which $\omega_k$ occurs in any of its reference sentences.

Our \cidern~score for \ngrams~of length $n$ is computed using the average cosine similarity between the candidate sentence and the reference sentences, which accounts for both precision and recall:
\begin{equation}\label{eq:2}
\mathcider_n(c_i, \RS_i) = \frac{1}{m}\sum_j \frac{\bm{g^n}(c_{i})\cdot \bm{g^n}(\rs_{ij})}{\|\bm{g^n}(c_{i})\|\|\bm{g^n}(\rs_{ij})\|},
\end{equation}
where $\bm{g^n}(c_{i})$ is a vector formed by $g_k(c_{i})$ corresponding to all \ngrams\ of length $n$ and $\|\bm{g^n}(c_{i})\|$ is the magnitude of the vector $\bm{g^n}(c_{i})$. Similarly for $\bm{g^n}(\rs_{ij})$.

%Our metric captures \ngram~level saliency via the IDF weighting, and accuracy-based measure  compute an accuracy-based measure for each sentence and take a mean across references to capture the consensus in descriptions.

We use higher order (longer) \ngrams~to capture grammatical properties as well as richer semantics. We combine the scores from \ngrams~of varying lengths as follows:

%We vary the weighting of $\mathcider_n$ scores computed using different size \ngrams:
\begin{equation} \label{eq:3}
\mathcider(c_i, \RS_i) = \sum_{n=1}^N w_n \mathcider_n(c_i, \RS_i),
\end{equation}
Empirically, we found that uniform weights $w_n=1/N$ work the best. We use $N$ = 4.

%\devi{Simplicity of an evaluation metric is critical to ensure that it is interpretable, transparent and easily adopted.}
%We also experimented with non-uniform $w_n$ that weighs the longer \ngrams more, soft word-level semantic similarity measures such as WordNet similarity~\cite{Miller:1995:WLD:219717.219748} and word2vec similarity~\cite{mikolov2013distributed}, as well as various relevance weighting schemes from information retrieval~\cite{Robertson04understandinginverse}. We found that the simple metric we present here performs the best. Interestingly, we find that word-level semantic similarity helps at fewer number of sentences. These soft-similarities help overcome the lack of data. But with more sentences, the noise introduced by existing similarity measures holds the metric back. Moreover, simplicity is critical to ensure that the metric is interpretable, transparent, and easy to adopt.

\section{New Datasets}
\label{sec:dataset}
We propose two new datasets -- \pascal and \abstractscenes~-- for evaluating image caption generation methods. Both the datasets have 50 reference sentences per image for 1,000 and 500 images respectively. These are intended as ``testing" datasets, crafted to enable consensus-based evaluation. For a list of training datasets, we encourage the reader to explore~\cite{LinECCV14coco,Ordonez:2011:im2text}. The \pascal dataset uses all 1,000 images from the UIUC Pascal Sentence Dataset~\cite{Rashtchian:2010:CIA:1866696.1866717} whereas the \abstractscenes dataset uses 500 random images from the Abstract Scenes Dataset~\cite{ZitnickCVPR2013}. The Abstract Scenes Dataset contains scenes made from clipart objects. Our two new datasets are different from each other both visually and in the type of image descriptions produced.

Our goal was to collect image descriptions that are objective and representative of the image content. Subjects were shown an image and a text box, and were asked to ``Describe what is going on in the image''. We asked subjects to capture the main aspects of the scene and provide descriptions that others are also likely to provide. This includes writing descriptions rather than ``dialogs" or overly descriptive sentences. 
%They were encouraged to capture the main aspects of the scene. 
Workers were told that a good description should help others recognize the image from a collection of similar images. Instructions also mentioned that work with poor grammar would be rejected. Snapshots of our interface can be found in the appendix. Overall, we had 465 subjects for \abstractscenes and 683 subjects for \pascal datasets. We ensure that each sentence for an image is written by a different subject. The average sentence length for the \abstractscenes dataset is 10.59 words compared to 8.8 words for \pascal. 
%\rama{Removing this sentence: This is indicative of a denser semantic sampling in the \abstractscenes dataset as compared to the \pascal dataset.}
\section{Experimental Setup}
\label{sec:experiments}
The goals of our experiments are two-fold:
%three-fold:
\begin{packed_itemize}
\item Evaluating how well our proposed metric \cider~captures human judgement of consensus, as compared to existing metrics.
\item Comparing existing state-of-the-art automatic image description approaches in terms of how well the descriptions they produce match human consensus of image descriptions.
%\item Qualitatively analyzing images from different datasets in terms of the level of consensus in the different human sentences describing the images.
\end{packed_itemize}

We first describe how we select candidate sentences for evaluation and the metrics we use for comparison to \cider. Finally, we list the various automatic image description approaches and our experimental set up.

%Finally, we discuss our setup for analyzing the level of consensus in human sentences across various datasets.
%\devi{Finally, $\ldots$} \pdfcomment{the way I've written this, feels like it might be ok to add in the experimental set up info for the qualitative consensus COCO, etc. stuff?}

%we describe our approach to evaluating the existing image description methods against each other. Finally, we describe the existing metrics used for evaluating image descriptions. More details can be found in the supplementary material.

\vspace{-10pt}
\paragraph{Candidate Sentences:}
On \abstractscenes, we use 48 of our 50 sentences as reference sentences (sentence A in our triplet annotation). The remaining 2 sentences per image can be used as candidate sentences. We form 400 pairs of candidate sentences (B and C in our triplet annotation). These include two kinds of pairs. The first are 200 human--human correct pairs (HC), where we pick two human sentences describing the same image. The second kind are 200  human--human incorrect pairs (HI), where one of the sentences is a human description for the image and the other is also a human sentence but describing some other image from the dataset picked at random.%\rama{Do you still think candidate sentences should be another section?}

For \pascal, our candidate sentences come from a diverse set of sources: human sentences from the UIUC Pascal Sentence Dataset as well as machine-generated sentences from five automatic image description methods. These span both retrieval-based and generation-based methods: Midge~\cite{midge}, Babytalk~\cite{babytalk}, Story~\cite{Story}, and two versions of Translating Video Content to Natural Language Descriptions~\cite{rohrbach13iccv} (Video and Video+).\footnote{We thank the authors of these approaches for making their outputs available to us.} We form 4,000 pairs of candidate sentences (again, B and C for our triplet annotation). These include four types of pairs (1,000 each). The first two are human--human correct (HC) and human--human incorrect (HI) similar to \abstractscenes. The third are human--machine (HM) pairs formed by pairing a human sentence describing an image with a machine generated sentence describing the same image. Finally, the fourth are machine--machine (MM) pairs, where we compare two machine generated sentences describing the same image. We pick the machine generated sentences randomly, so that each method participates in roughly equal number of pairs, on a diverse set of images. Ours is the first work to perform a comprehensive evaluation across these different kinds of sentences.

For consistency, we drop two reference sentences for the \pascal evaluations so that we evaluate on both datasets (\abstractscenes and \pascal) with a maximum of 48 reference sentences.

\vspace{-10pt}
\paragraph{Metrics:}
The existing metrics used in the community for evaluation of image description approaches are \bleu~\cite{papineni}, \rouge~\cite{rouge} and \meteor~\cite{meteor}. \bleu is precision-based and \rouge is recall-based. More specifically, image description methods have used versions of \bleu called \bleux$_1$ and \bleux$_4$, and a version of \rouge called \rougex$_1$. A recent survey paper~\cite{elliott-keller:2014:P14-2} has used a different version of \rouge called \rougex$_S$, as well as the machine translation metric called \meteor~\cite{meteor}. We now briefly describe these metrics. More details can be found in the appendix.
%We perform a procedure called stemming ~\cite{add}, which matches word forms to a common root. For instance ``plays", ``played" and ``playing" get mapped to a common word. This preprocessing is done for all metrics except \meteor, which does its own pre-processing.
%\begin{itemize}
%\item
{\textbf \bleu} (BiLingual Evaluation Understudy)~\cite{papineni} is a popular metric for Machine Translation (MT) evaluation. It computes an \ngram based precision for the candidate sentence with respect to the references. The key idea of \bleu is to compute precision by \emph{clipping}. Clipping computes precision for a word, based on the maximum number of times it occurs in any reference sentence. Thus, a candidate sentence saying ``The The The", would get credit for saying only one ``The", if the word occurs at most once across individual references. BLEU computes the geometric mean of the n-gram precisions and adds a “brevity-penalty” to discourage overly short sentences. The most common formulation of BLEU is BLEU4, which uses 1-grams up to 4-grams, though lower-order variations such as BLEU1 (unigram BLEU) and BLEU2 (unigram and bigram BLEU) are also used. Similar to~\cite{elliott-keller:2014:P14-2,journals/jair/HodoshYH13} for evaluating image descriptions, we compute BLEU at the sentence level. For machine translation BLEU is most often computed at the corpus level where correlation with human judgment is high; the correlation is poor at the level of individual sentences. In this paper we are specifically interested in the evaluation of accuracies on individual sentences.
%\item
{\textbf \rouge} stands for Recall Oriented Understudy of Gisting Evaluation~\cite{rouge}. It computes \ngram based recall for the candidate sentence with respect to the references. It is a popular metric for summarization evaluation. Similar to \bleu, versions of \rouge can be computed by varying the \ngram count. Two other versions of \rouge are \rougex$_S$ and \rougex$_L$. These compute an F-measure with a recall bias using \emph{skip-bigrams} and \emph{longest common subsequence} respectively, between the candidate and each reference sentence. Skip-bigrams are all pairs of ordered words in a sentence, sampled non-consecutively. Given these scores, they return the maximum score across the set of references as the judgment of quality.
%\item
{\textbf \meteor} stands for Metric for Evaluation of Translation with Explicit ORdering~\cite{meteor}. Similar to \rougex$_L$ and \rougex$_S$, it also computes the F-measure based on matches, and returns the maximum score over a set of references as its judgment of quality. However, it resolves word-level correspondences in a more sophisticated manner, using exact matches, stemming and semantic similarity. It optimizes over matches minimizing \emph{chunkiness}. Minimizing chunkiness implies that matches should be consecutive, wherever possible. It also sets parameters favoring recall over precision in its F-measure computation. We implement all the metrics, except for \meteor, for which we use~\cite{denkowski:lavie:meteor-wmt:2014} (version 1.5). Similar to \bleu, we also aggregate \meteor scores at the sentence level.
%\end{itemize}

\vspace{-10pt}
\paragraph{Machine Approaches:}
We comprehensively evaluate which machine generation methods are best at matching consensus sentences. For this experiment, we select a subset of 100 images from the UIUC Pascal Sentence Dataset for which we have outputs for all the five machine description methods used in our evaluation: Midge~\cite{midge}, Babytalk~\cite{babytalk}, Story~\cite{Story}, and two versions of Translating Video Content to Natural Language Descriptions~\cite{rohrbach13iccv} (Video and Video+). For each image, we form all $\comb{5}{2}$ pairs of machine--machine sentences. This ensures that each machine approach gets compared to all other machine approaches on each image. This gives us 1,000 pairs. We form triplets by ``tripling'' each pair with 20 random reference sentences. We collect human judgement of consensus using our triplet annotation modality as well as evaluate our proposed automatic consensus metric \cider using the same reference sentences. In both cases, we count the fraction of times a machine description method beats another method in terms of being more similar to the reference sentences. To the best of our knowledge, we are the first work to perform an exhaustive evaluation of automated image captioning, across retrieval- and generation-based methods. %\rama{100 is around 35 \% of all data. I think this should be fine, these methods have a distinct quality in terms of sentences they produce, 100 should be enough to give a good signal}.

%\paragraph{Consensus in Image Descriptions}
%We evaluate which images have more consensus in descriptions on the \abstractscenes, \pascal and the newly released \coco dataset, containing 5 descriptions per image~\cite{}. We use the full datasets for the first two, and sample 30k out of 110k \coco images. The consensus evaluation is performed by computing the \cider score in a leave one out fashion for reference sentences. That is, for a given image, one sentence is treated as the candidate sentence and its score is computed with respect to the rest of the reference sentences. This is repeated $N$ times where $N$ is the number of sentences per image. We call images with higher mean \cider scores as having more consensus and vice versa. 
\section{Results}
\label{sec:results}
In this section we evaluate the effectiveness of our consensus-based metric \cider on the \pascal and \abstractscenes datasets. We begin by exploring how many sentences are sufficient for reliably evaluating our consensus metric. Next, we compare our metric against several other commonly used metrics on the task of matching human consensus. Then, using \cider we evaluate several existing automatic image description approaches. Finally, we compare performance of humans and \cider at predicting consensus.

\subsection{How many sentences are enough?}
%To answer this question,
We begin by analyzing how the number of reference sentences affects the accuracy of automated metrics. To quantify this, we collect 120 sentences for a subset of 50 randomly sampled images from the UIUC Pascal Sentence Dataset. We then pool human--human correct, human--machine, machine--machine and human--human incorrect sentence pairs (179 in total) and get triplet annotations. This gives us the ground truth consensus score for all pairs.
%For each automated metric, we evaluate with the ``best" collected triplet ground truth (at 48 sentences).
We evaluate \bleux$_1$, \rougex$_1$ and \ciderx$_1$ with up to 100 reference sentences used to score the candidate sentences. We find that the accuracy improves for the first 10 sentences (Fig. \ref{fig:3}) for all metrics. From 1 to 5 sentences, the agreement for \rougex$_1$ improves from 0.63 to 0.77. Both \rougex$_1$ and \ciderx$_1$ continue to improve until reaching 50 sentences, after which the results begin to saturate somewhat. Curiously, \bleux$_1$ shows a decrease in performance with more sentences. \bleu does a max operation over sentence level matches, and thus as more sentences are used, the likelihood of matching a lower quality reference sentence increases. Based on this pilot, we collect 50 sentences per image for our \abstractscenes and \pascal datasets. For the remaining experiments we report results using 1 to 50 sentences.

\begin{figure*}
\centering
\begin{subfigure}{0.33\textwidth}
\includegraphics[width=\textwidth]{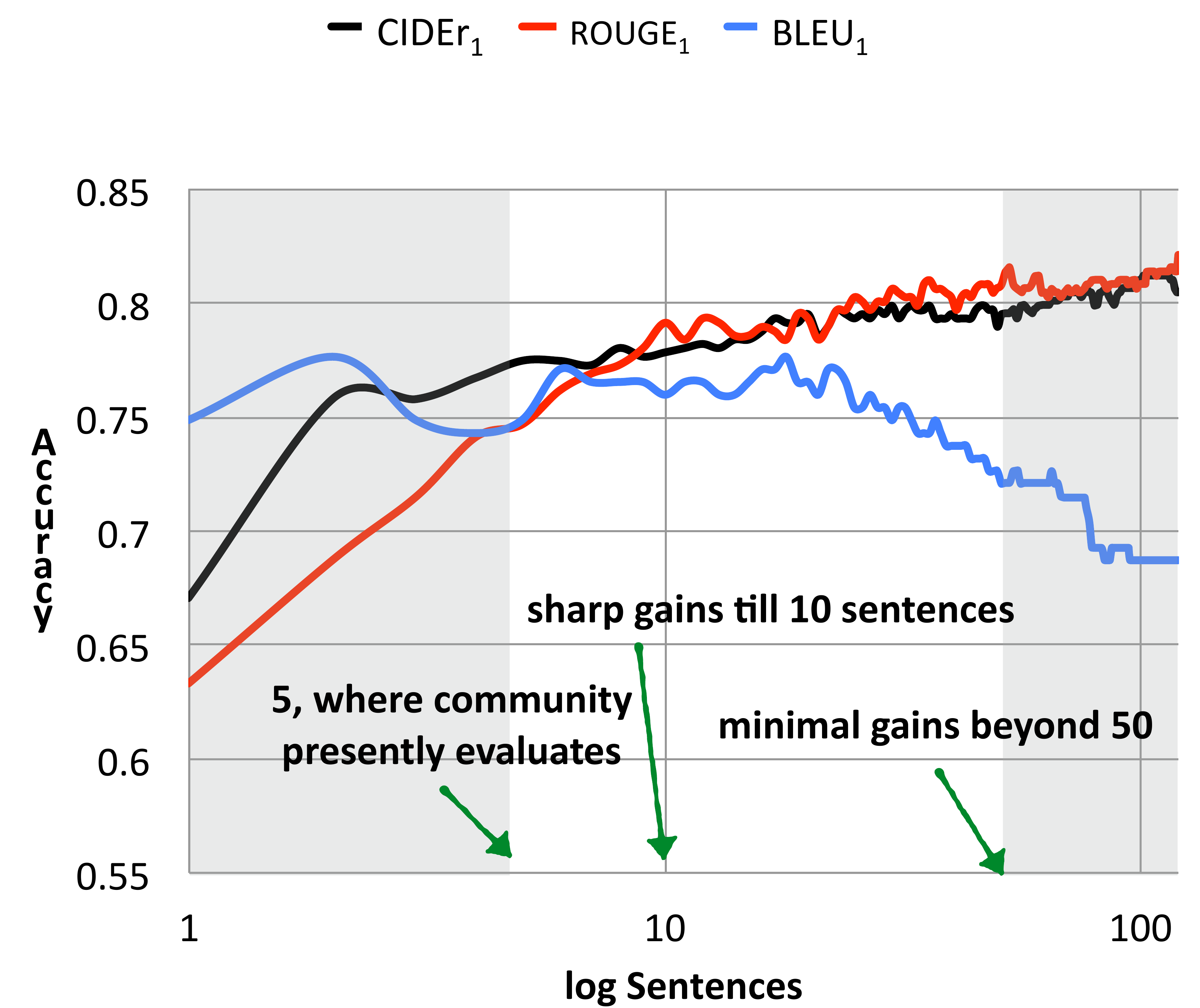}
\caption{Pilot Performance}
\vspace{-10pt}
\label{fig:3}
\end{subfigure}
\begin{subfigure}{0.33\textwidth}
\includegraphics[width=\textwidth, page=1]{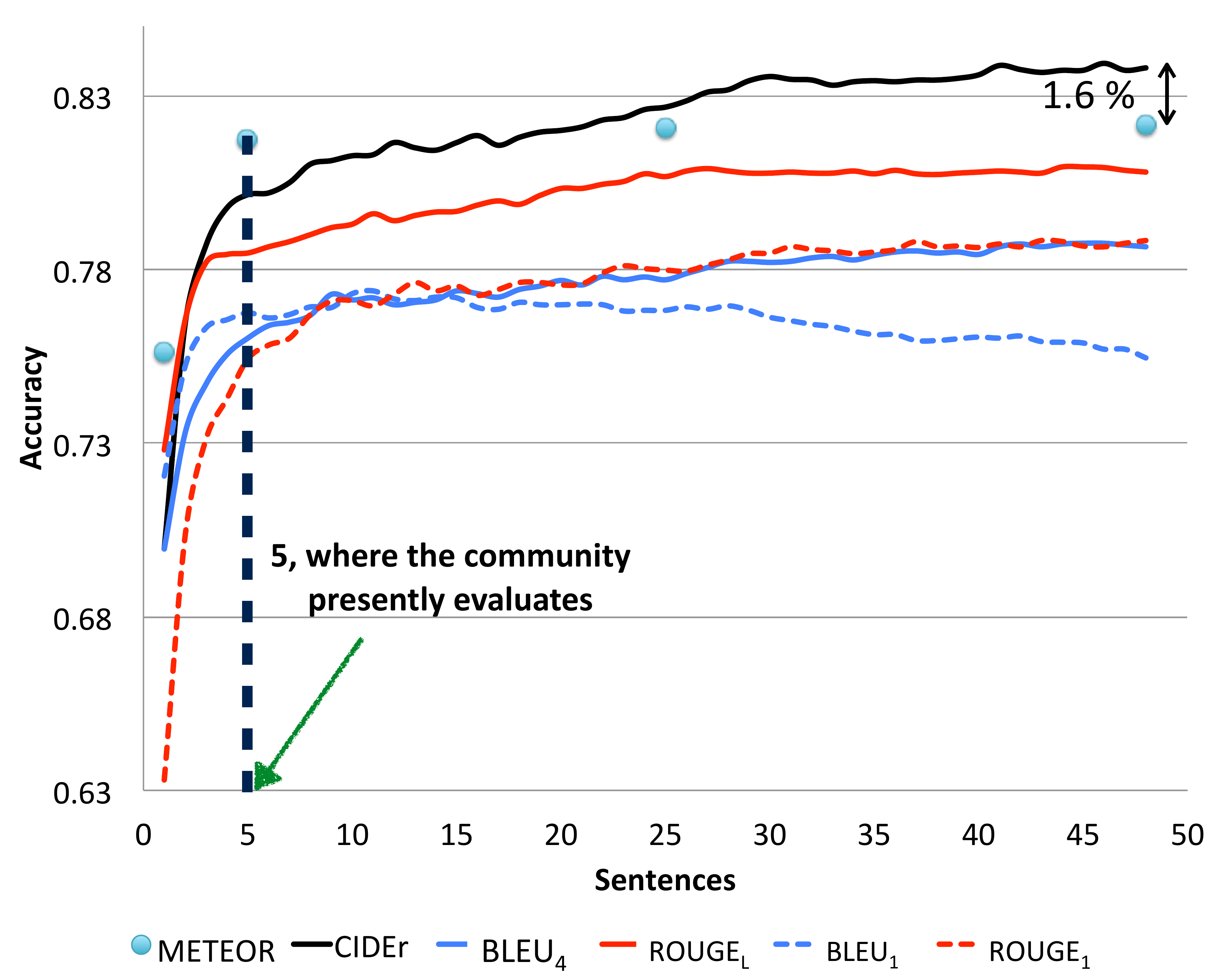}
\caption{\pascal}
\end{subfigure}
\begin{subfigure}{0.33\textwidth}
\includegraphics[width=\textwidth, page=2]{figures/main}
\caption{\abstractscenes}
\end{subfigure}
\caption{(\textbf{a}): We show accuracy (y-axis) versus \emph{log} number of sentences (x-axis) for our pilot study. We note that the gains saturate after 50 sentences. \textbf{(b) and (c):} Accuracy of automated metrics (y-axis) plotted against number of reference sentences (x-axis) for \pascal (b) and \abstractscenes (c). Metrics currently used for evaluating image descriptions are shown in \emph{dashed} lines. Other existing metrics and our proposed metric are in \textbf{solid} lines. \cider is the best performing metric on both datasets followed by \meteor. \meteor is sampled at fewer points, due to high run-time. Note that more reference sentences that we collect clearly help.
}
\vspace{-10pt}
\label{fig:4}
\end{figure*}

\subsection{Accuracy of Automated Metrics}
We evaluate the performance of \cider, \bleu, \rouge and \meteor at matching the human consensus scores in Fig. \ref{fig:4}. That is, for each metric we compute the scores for two candidate sentences. The metric is correct if the sentence with higher score is the same as the sentence chosen by our human studies as being more similar to the reference sentences. The candidate sentences are both human and machine generated. For \bleu and \rouge we show both their popular versions and the version we found to give best performance. We sample \meteor at fewer points due to high run-time. For a more comprehensive evaluation across different versions of each metric, please see the appendix.

At 48 sentences, we find that \cider is the best performing metric, on both \abstractscenes as well as \pascal. It is followed by \meteor on each dataset. Even using only 5 sentences, both \cider and \meteor perform well in comparison to \bleu and \rouge. \cider beats \meteor at 5 sentences on \abstractscenes, whereas \meteor does better at five sentences on \pascal. This is because \meteor incorporates soft-similarity, which helps when using fewer sentences. However, \meteor, despite its sophistication does a max across reference scores, which limits its ability to utilize larger numbers of reference sentences. Popular metrics like \rougex$_1$ and \bleux$_1$ are not as good at capturing consensus.  \cider provides consistent performance across both the datasets, giving 84\% and 84\% accuracy on \pascal and \abstractscenes respectively.
%In contrast \meteor drops by  1.4\% in accuracy on the semantically dense \abstractscenes dataset.

Considering previous papers only used 5 reference sentences per image for evaluation, the relative boost in performance is substantial. Using \bleux$_1$ or \rougex$_1$ at 5 sentences, we obtained 76\% and 74\% accuracy on \pascal. With \cider at 48 sentences, we achieve 84\% accuracy. This brings automated evaluation much closer to human performance (90\%, details in Sec.~\ref{sec:inter-human}). On the Flickr8K dataset~\cite{journals/jair/HodoshYH13} with human judgments on 1-5 ratings, \meteor has a correlation (Spearman's $\rho$) of 0.56 \cite{elliott-keller:2014:P14-2}, whereas \cider achieves a correlation of 0.58 with human judgments.\footnote{We thank Desmond Elliot for the result.}

We next show the best performing versions of the metrics \cider, \bleu, \rouge and \meteor on \pascal and \abstractscenes, respectively, for different kinds of candidate pairs (Table~\ref{table:1}). As discussed in Sec.~\ref{sec:dataset} we have four kinds of pairs: (human--human correct) HC, (human--human incorrect) HI, (human--machine) HM, and (machine--machine) MM. We find that out of six cases, our proposed automated metric is best in five. We show significant gains on the challenging MM and HC tasks that involve differentiating between fine-grained differences between sentences (two machine generated sentences and two human generated sentences). This result is encouraging because it indicates that the \cider metric will continue to perform well as image description methods continue to improve. On the easier tasks of judging consensus on HI and HM pairs, all methods perform well.
\begin{table}
\begin{center}
\begin{tabular}{|c|c|c|c|c|c|c|}
\hline
\multicolumn{1}{|c|}{Metric} & \multicolumn{4}{c|}{\footnotesize{\pascal}} & \multicolumn{2}{c|}{\footnotesize{\abstractscenes}}\\
\hline
& HC & HI & HM & MM & HC & HI\\
\hline
\bleux$_4$ & 64.8  & 97.7 & 93.8 & 63.6 & 65.5 & 93.0\\
\rouge & 66.3  & 98.5 & 95.8 & 64.4 & \textbf{71.5} & 91.0\\
\meteor & 65.2  & 99.3 & \textbf{96.4} & 67.7 & 69.5 & 94.0\\
\cider & \textbf{71.8} & \textbf{99.7} & 92.1 & \textbf{72.2} & \textbf{71.5} & \textbf{96.0}\\
\hline
\end{tabular}
\end{center}
\vspace{-10pt}
\caption{Results on four kinds of pairs for \pascal and two kinds of pairs for \abstractscenes. The best performing method is shown in \textbf{bold}. Note: we use \rougex$_L$ for \pascal and \rougex$_1$ for \abstractscenes}
\vspace{-10pt}
\label{table:1}
\end{table}

\subsection{Which automatic image description approaches produce consensus descriptions?}
\label{sec:benchmark}
\begin{figure}[tbp]
\includegraphics[width=\columnwidth, page=1]{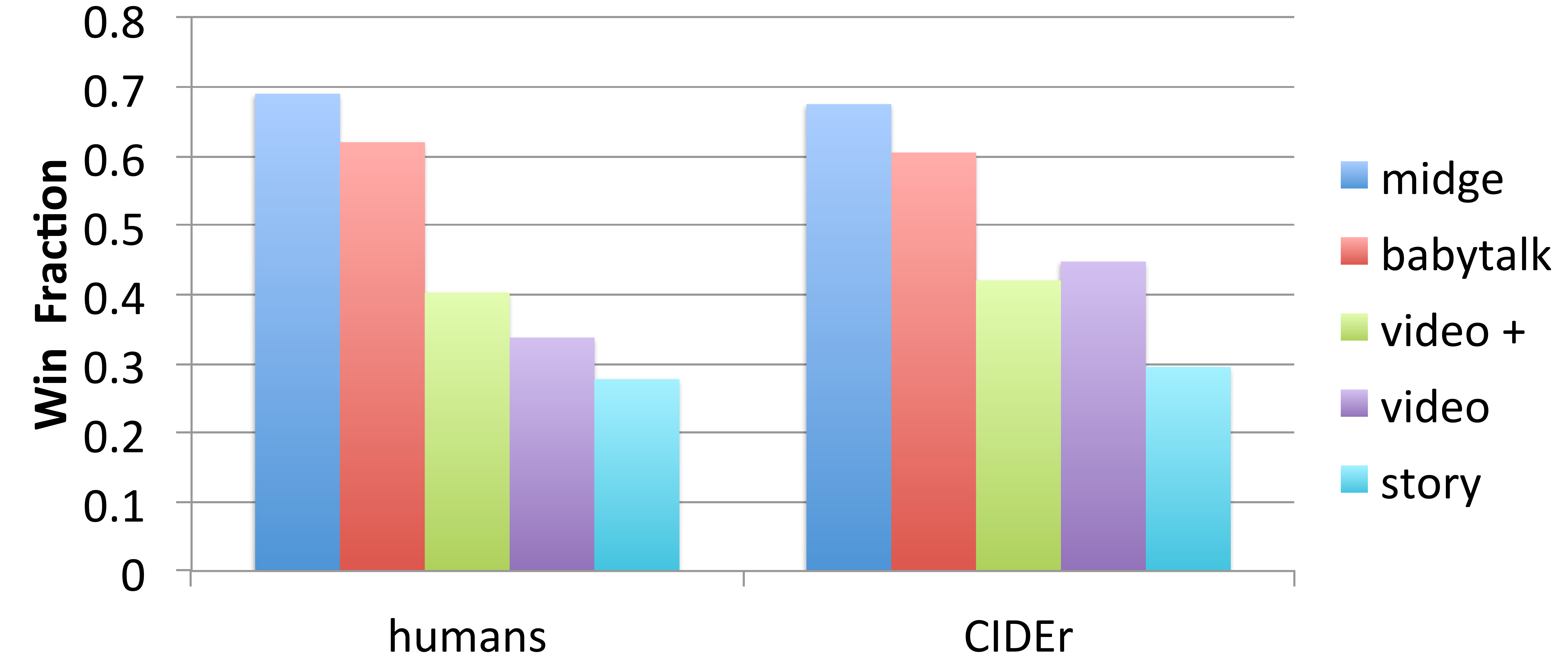}
\caption{Fraction of times a machine generation approach wins against the other four (y-axis), plotted for human annotations and our automated metric, \cider. }
\vspace{-10pt}
\label{fig:5}
\end{figure}
We have shown that \cider and our new datasets containing 50 sentences per image provide a more accurate metric over previous approaches. We now use it to evaluate some existing automatic image description approaches. Our methodology for conducting this experiment is described in Sec. \ref{sec:experiments}. Our results are shown in Fig. \ref{fig:5}. We show the fraction of times an approach is rated better than other approaches on the y-axis. We note that Midge~\cite{midge} is rated as having the best consensus by both humans and \cider, followed by Babytalk~\cite{babytalk}. Story~\cite{Story} is the lowest ranked, by both humans and \cider. Humans and \cider differ on the ranking of the two video approaches (Video and Video+)~\cite{rohrbach13iccv}. We calcuate the Pearson's correlation between the fraction of wins for a method on human annotations and using \cider. We find that humans and \cider agree with a high correlation (0.98).

%\begin{table}
%\begin{center}
%\begin{tabular}{|c|c|}
%\hline
%Method & \ciderx-D \\
%\hline
%M1 & 0.64  \\
%M2 & 0.66  \\
%M3 & 0.63  \\
%M4 & \textbf{.83} \\
%Human & .85 \\
%\hline
%\end{tabular}
%\end{center}
%\vspace{-10pt}
%\caption{\ciderx-D scores of very recent methods evaluated on \coco caption evaluation server.}
%\label{table:2}
%\vspace{-10pt}
%\end{table}

\subsection{Human Performance}\label{sec:inter-human}
In our final set of experiments we measure human performance at predicting which of two candidate sentences better matches the consensus.
%measured using our triplet annotation (Section \ref{sec:human_studies}).
Human performance puts into context how clearly consensus is defined, and provides a loose bound on how well we can expect automated metrics to perform. We evaluate both human and machine performance at predicting consensus on all 4,000 pairs from PASCAL-50S dataset and 400 pairs from the ABSTRACT-50S dataset described in Sec.~\ref{sec:experiments}. To create the same experimental set up for both humans and machines, we obtain ground truth consensus for each of the pairs using our triplet annotation on 24 references out of 48. For predicting consensus, humans (via triplet annotations) and machines both use the remaining 24 sentences as reference sentences.
% Thus, we must make automated metrics and humans comparable. We simulate a situation where both humans and machines make use of the same experimental settings. We
 %calculate the human ground truth using a fixed set of 24 reference sentences.
% Keeping this fixed, we pick multiple sets of 24 from the 48 references and evaluate humans and machines respectively to get predictions. These are then compared to annotations on the fixed set. Performance is averaged across three runs.
 We find that the best machine performance is 82\% on \pascal using \cider, in contrast to human performance which is at 90\%. On the \abstractscenes dataset, \cider is at 82\% accuracy, whereas human performance is at 83\%.

\section{Gameability and Evaluation Server}
%In this work we propose a novel consensus based evaluation protocol for image descriptions. We introduce an interpretable, transparent and easy to adopt metric, that performs well on consensus based evaluation.
\paragraph{Gameability}\label{para:ciderd}
When optimizing an algorithm for a specific metric undesirable results may be achieved. The ``gaming'' of a metric may result in sentences with high scores, yet produce poor results when judged by a human. To help defend against the future gaming of the CIDEr metric, we propose several modifications to the basic CIDEr metric called CIDEr-D.

First, we propose the removal of stemming. When performing stemming the singular and plural forms of nouns and different tenses of verbs are mapped to the same token. The removal of stemming ensures the correct forms of words are used. Second, in some cases the basic CIDEr metric produces higher scores when words of higher confidence are repeated over long sentences. To reduce this effect, we introduce a Gaussian penalty based on the difference between candidate and reference sentence lengths. Finally, the sentence length penalty may be gamed by repeating confident words or phrases until the desired sentence length is achieved. We combat this by adding clipping to the \ngram counts in the \cidern numerator. That is, for a specific \ngram we clip the number of candidate occurrences to the number of reference occurrences. This penalizes the repetition of specific \ngrams beyond the number of times they occur in the reference sentence. These changes result in the following equation (analogous to Equation~\ref{eq:2}):

\begin{multline}
\mathciderD_n(c_i, \RS_i) = \frac{10}{m}\sum_j e^{\frac{-(l(c_{i})-l(\rs_{ij}))^2}{2\sigma^2}} * \\ \frac{\text{min}(\bm{g^n}(c_{i}),\bm{g^n}(\rs_{ij})) \cdot \bm{g^n}(\rs_{ij})}{\|\bm{g^n}(c_{i})\|\|\bm{g^n}(\rs_{ij})\|},
\end{multline}

Where $l(c_{i})$ and $l(\rs_{ij})$ denote the lengths of candidate and reference sentences respectively. We use $\sigma = 6$. A factor of 10 is added to make the \ciderx-D scores numerically similar to other metrics.

The final CIDEr-D metric is computed in a similar manner to CIDEr (analogous to Equation~\ref{eq:3}):
\begin{equation}
\mathciderD(c_i, \RS_i) = \sum_{n=1}^N w_n \mathciderD_n(c_i, \RS_i),
\end{equation}
Similar to CIDEr, uniform weights are used. We found that this version of the metric has a rank correlation (Spearman's $\rho$) of 0.94 with the original CIDEr metric while being more robust to gaming. Qualitative examples of ranking can be found in the appendix.

\paragraph{Evaluation Server} To enable systematic evaluation and benchmarking of image description approaches based on consensus, we have made \ciderx-D available as a metric in the \coco caption evaluation server~\cite{2015arXiv150400325C}.

\section{Conclusion}
In this work we proposed a consensus-based evaluation protocol for image description evaluation. Our protocol enables an objective comparison of machine generation approaches based on their ``human-likeness", without having to make arbitrary calls on weighing content, grammar, saliency, \etc. with respect to each other. We introduce an annotation modality for measuring consensus, a metric \cider for automatically computing consensus, and two datasets, \pascal and \abstractscenes with 50 sentences per image. We demonstrate \cider has improved accuracy over existing metrics for measuring consensus. 
\paragraph{Acknowledgements:}We thank Chris Quirk, Margaret Mitchell and Michel Galley for helpful discussions in formulating \ciderx-D. This work was supported in part by The Paul G. Allen Family Foundation Allen Distinguished Investigator award  to D.P.
%Our results demonstrate that our proposed automated metric, \cider, outperforms performs really well at maching consensus, and generalizes across our two new datasets, \pascal and \abstractscenes. 

\clearpage
{\small
\bibliographystyle{ieee}
\bibliography{consensus_cvpr_camera_ready}
}
\clearpage

\section*{Appendix Overview}
List of items:
\begin{enumerate}[label=\Roman*]
\item \textbf{Comparison of metrics on triplet annotations to pairwise annotations:} Compares the accuracy of \cider on triplet annotation to existing choices of metrics on pairwise annotations
\item \textbf{Ranking of reference sentences for various automated metrics:} Qualitative examples of the kind of sentences preferred by each metric
\item \textbf{Comparison of rankings from \cider and \ciderx-D:} Establishes that both \cider and \ciderx-D are similar qualitatively, in terms of how they rank reference sentences
\item \textbf{Difference between human-like and what humans like:} Shows examples of differences between pairwise and triplet annotations. Pairwise annotations often favor longer sentences
\item \textbf{Sentence collection interface for \pascal and \abstractscenes:} Shows a snapshot of the interface used to collect our datasets, and explains the instructions
\item \textbf{Equations for \bleu, \rouge, and \meteor:} Formulates some existing metrics in terms of the notation used in the rest of the paper
\item \textbf{Qualitative examples of outputs of image description methods evaluated in the paper:} Gives a sense for the kind of outputs produced by each of the image description methods evaluated in the paper
\item \textbf{Performance of different versions of metrics on consensus:} Benchmarks the performance of different versions of metrics discussed in the paper at matching human consensus 
\end{enumerate}

\subsection*{Appendix I : Comparison to Pairwise Annotations}
\label{sec:1}
We consider some alternate annotation modalities and compare the performance of present metrics on them with that of \cider on consensus. The first such modality is a pairwise interface described as follows. Subjects on Amazon Mechanical Turk (AMT) are shown just the two candidate sentences (B and C) with the image (instead of sentence A), and asked to pick the \emph{better} description out of the two. 11 such human judgments are collected for each such pair. These annotations are collected for the same \pascal candidate sentences as those used for the triplet experiments in the paper. We compare accuracy on \emph{consensus} for \cider to accuracy of other metrics on picking the  \emph{better} candidate sentence.  We find that \rougex$_L$ at 5 sentences performs at 75.6\% whereas the \bleux$_4$ version performs at 74.75\%. \rougex$_1$ and \bleux$_1$ perform at 73.15\% and 73.4\% respectively at 5 sentences. With \meteor at 5 sentences, the performance is at 79.5\%. In contrast, \cider at 48 sentences reaches an accuracy of 84\% on consensus. Thus the consensus-based protocol comprising of our proposed metric, dataset and human annotation modality provides more accurate automated evaluation. 
\subsection*{Appendix II : Ranking of Sentences}
\label{sec:2}
\begin{figure*}[t]
\centering
\includegraphics[scale = 0.44, page=3]{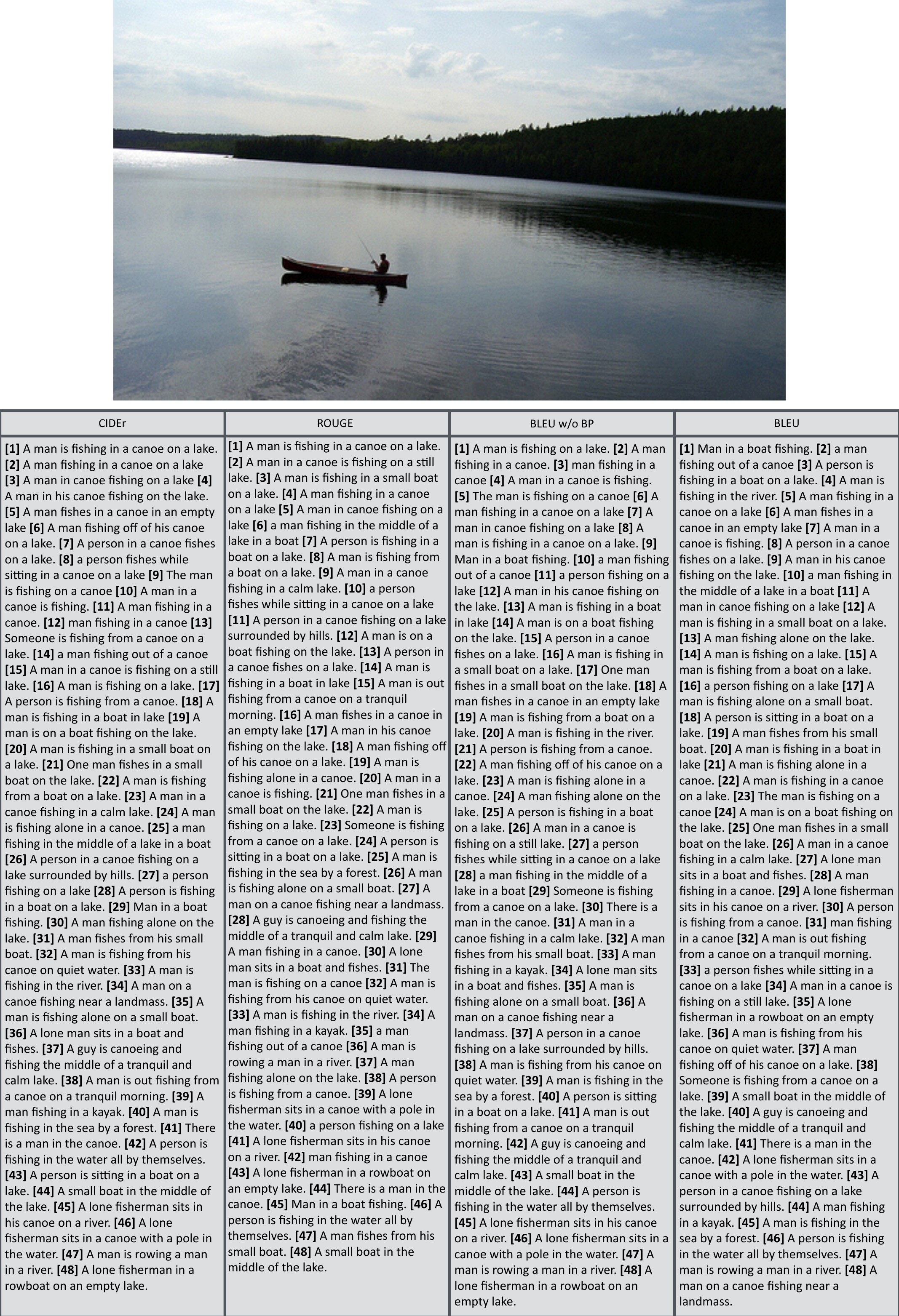}
\caption{Ranking of 48 sentences, from highest score to lowest score, as predicted by each metric. Notice how \cider captures how most humans tend to describe an image (consensus) better, wheareas \rouge scores invariably favor longer, detailed sentences (less salient) and \bleu scores favor shorter sentences (lacking coverage) when used without Brevity Penalty. \rougex$_1$ and \bleux$_1$ versions of \rouge and \bleu are used.}
\label{fig:p3}
\end{figure*}

We now show a ranking of the 48 sentences collected for a particular image as per the \cider, \bleux$_1$, \bleux$_1$ without Brevity Penalty and \rougex$_1$ scores (Fig.~\ref{fig:p3}). Each reference sentence is considered in turn as a candidate and scored with the remaining (47) reference sentences using the corresponding metric. Note how the top-ranked \cider sentences show high consensus. The top-ranked \rouge sentences are typically more detailed, whereas the top ranked \bleu sentences are not as consistent as those with \cider. If \bleu was used without the brevity penalty, as some previous works have~\cite{babytalk,Ordonez:2011:im2text} one would see that really short sentences get high scores. Intuitively, we can see that the ranking produced by \cider is more meaningful.

%\begin{figure*}[t]
%\centering
%\includegraphics[scale = 0.44, page=1]{figures/ranking_cr_examples}
%\caption{Ranking of 48 sentences, from highest score to lowest score, as predicted by each metric. Notice how \cider captures how most humans tend to describe an image (consensus) better, wheareas \rouge scores invariably favor longer, detailed sentences (less salient) and \bleu scores favor shorter sentences (lacking coverage) when used without Brevity Penalty. \rougex$_1$ and \bleux$_1$ versions of \rouge and \bleu are used.}
%\label{fig:p1}
%\end{figure*}

%\begin{figure*}[t]
%\centering
%\includegraphics[scale = 0.44, page=2]{figures/ranking_cr_examples}
%\caption{Ranking of 48 sentences, from highest score to lowest score, as predicted by each metric. Notice how \cider captures how most humans tend to describe an image (consensus) better, wheareas \rouge scores invariably favor longer, detailed sentences (less salient) and \bleu scores favor shorter sentences (lacking coverage) when used without Brevity Penalty. \rougex$_1$ and \bleux$_1$ versions of \rouge and \bleu are used.}
%\label{fig:p2}
%\end{figure*}

%\begin{figure*}[t]
%\includegraphics[width = \textwidth, page=4]{figures/ranking_examples}
%\caption{Ranking of 48 sentences, from highest score to lowest score, as predicted by each metric. Notice how \cider captures how most humans tend to describe an image (consensus) better, wheareas \rouge scores invariably favor longer, detailed sentences (less salient) and \bleu scores favor shorter sentences (lacking coverage). \rougex$_1$ and \bleux$_1$ without Brevity Penalty are used.}
%\label{fig:p4}
%\end{figure*}

%\begin{figure*}[t]
%\includegraphics[width = \textwidth, page=5]{figures/ranking_examples}
%\caption{Ranking of 48 sentences, from highest score to lowest score, as predicted by each metric. Notice how \cider captures how most humans tend to describe an image (consensus) better, wheareas \rouge scores invariably favor longer, detailed sentences (less salient) and \bleu scores favor shorter sentences (lacking coverage). \rougex$_1$ and \bleux$_1$ without Brevity Penalty are used. }
%\label{fig:p5}
%\end{figure*}

%\begin{figure*}[t]
%\includegraphics[width = \textwidth, page=6]{figures/ranking_examples}
%\caption{Ranking of 48 sentences, from highest score to lowest score, as predicted by each metric. Notice how \cider captures how most humans tend to describe an image (consensus) better, wheareas \rouge scores invariably favor longer, detailed sentences (less salient) and \bleu scores favor shorter sentences (lacking coverage). \rougex$_1$ and \bleux$_1$ without Brevity Penalty are used.}
%\label{fig:p6}
%\end{figure*}

\begin{figure*}[t]
\includegraphics[width = \textwidth, page=7]{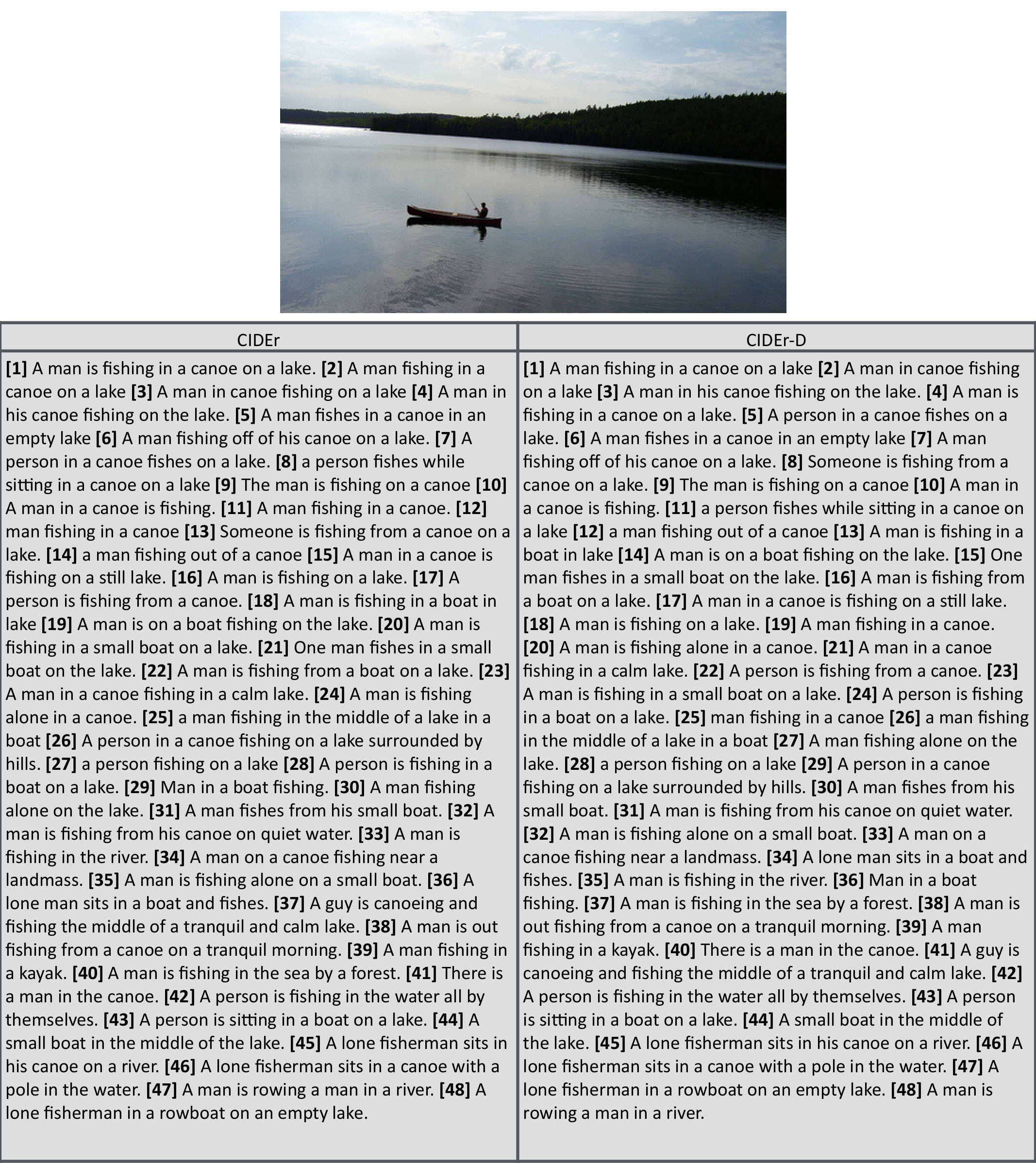}
\caption{Ranking of 48 sentences, from highest score to lowest score, as predicted by \ciderx$_1$ and \ciderx-D$_1$. Notice that the rankings are mostly similar qualitatively. \ciderx-D is more robust to gaming effects than \cider.}
\label{fig:p7}
\end{figure*}

\subsection*{Appendix III : Difference between Human-like and What Humans Like}
\label{sec:3}
In our experiments, we found that there can often be a difference in the sentence that is rated as ``better" (measured via pairwise annotation) by subjects \emph{versus} the kind of sentences written by subjects when asked to describe the image (measured via consensus annotation). We refer to this distinction as human-like vs what humans like. Some qualitative examples are shown in Fig.~\ref{fig:3}. Candidate sentences shown in bold are those that the consensus-based measure picks and those shown in thin font are those picked by the pairwise evaluation based on ``better". Reference sentences rated similar to the winning candidate sentence using the triplet annotation are shown in bold. 
\begin{figure*}[t]
\includegraphics[width=\textwidth]{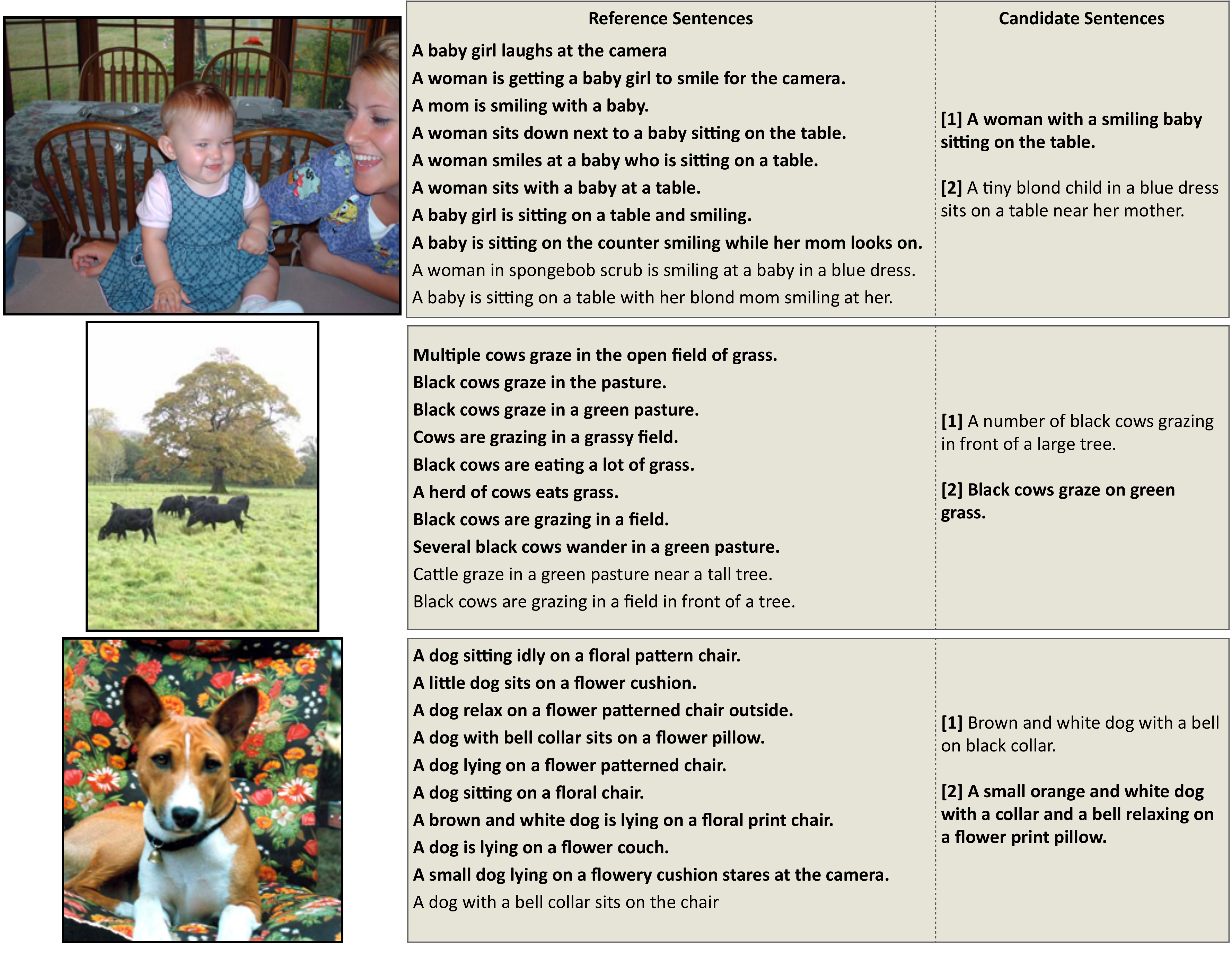}
\caption{Reference sentences shown in \textbf{bold} are those which are rated as more similar to the winning candidate sentence, also shown in \textbf{bold}, via the triplet interface. The candidate sentence not shown in bold is the one picked by the pairwise interface, which captures ``better". This illustrates the difference between human-like \emph{versus} what humans like.}
\label{fig:3}
\end{figure*}

\subsection*{Appendix IV : Ranking of sentences - \cider and \ciderx-D}
As we report in Sec.~\ref{para:ciderd}, we find that \cider and \ciderx-D agree with a high correlation (Spearman's $\rho$=0.94) on ranking of sentences. We now compare \cider$_1$ and \ciderx-D$_1$ rankings, since results are easier to interpret for the unigram case. An example of ranking can be found in Fig.~\ref{fig:p7}. Notice that the rankings of \cider and \ciderx-D are very similar qualitatively. However, the formulation of \ciderx-D avoids gaming effects as explained in Sec.~\ref{para:ciderd}.
\subsection*{Appendix V : Sentence Collection Interface}
\begin{figure}[t]
\includegraphics[width=\columnwidth]{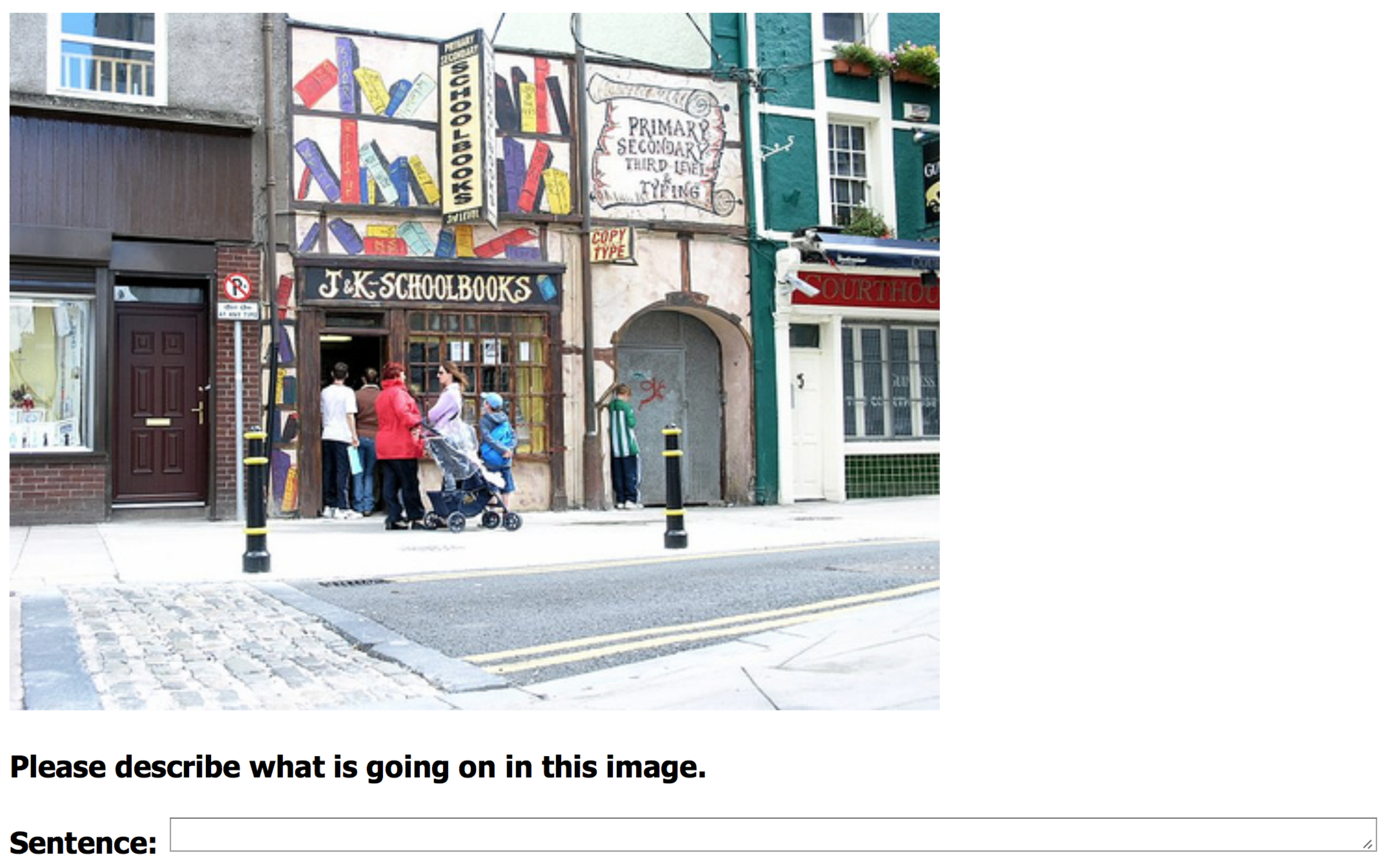}
\caption{Interface used for collecting image descriptions}
\label{fig:instructions}
\end{figure}
\begin{figure}[t]
\includegraphics[width=\columnwidth]{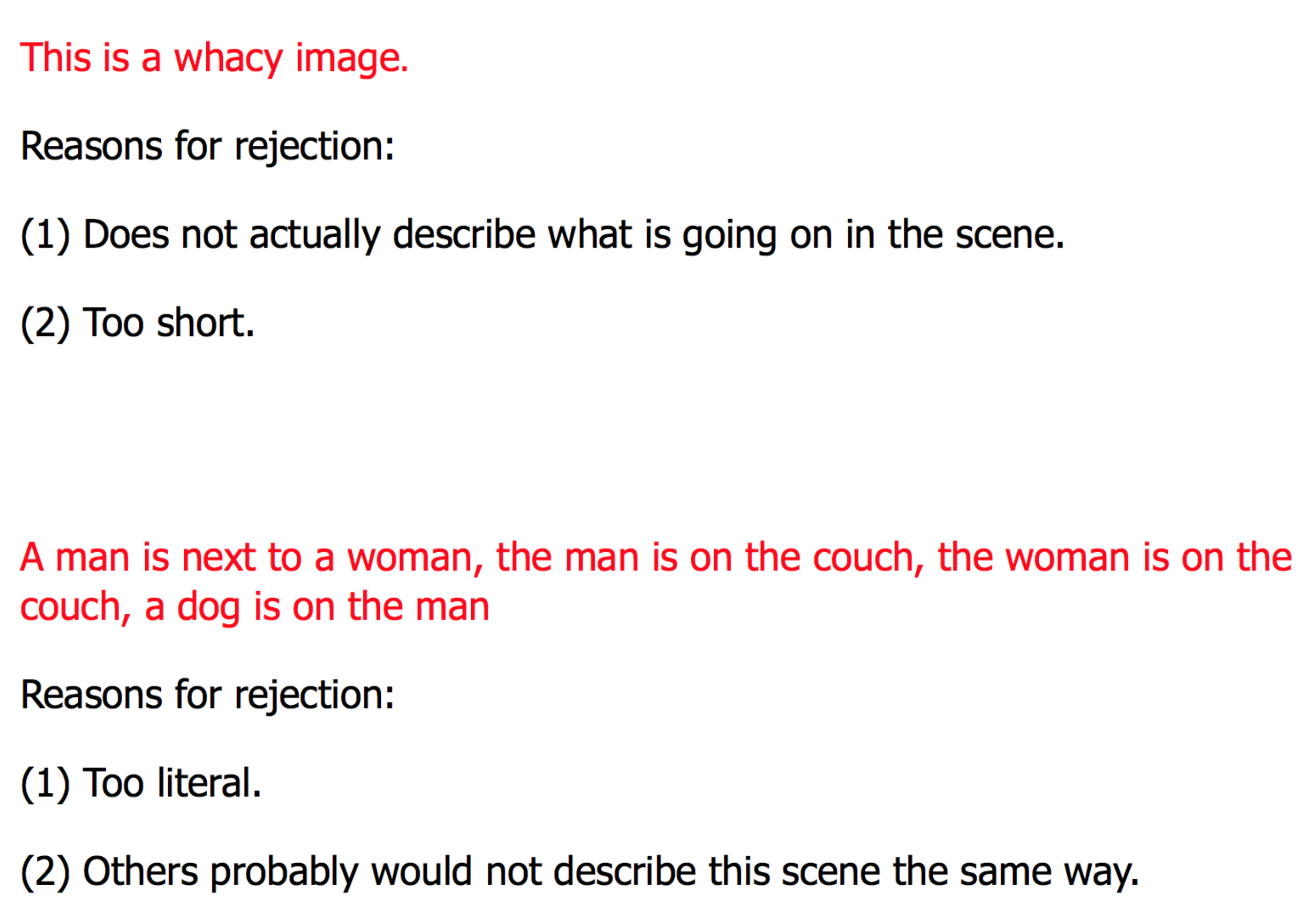}
\caption{An illustration of our rejection criteria with examples shown to subjects on Amazon Mechanical Turk (AMT)}
\label{fig:rejection}
\end{figure}
\label{sec:4}

The sentence collection interface for both \abstractscenes and \pascal is shown in Fig.~\ref{fig:instructions}. Stringent rejection criteria were specified (Fig.~\ref{fig:rejection}).
\subsection*{Appendix VI : Image Description Method Outputs}
\label{sec:5}
\begin{figure*}[t]
\includegraphics[width=\textwidth]{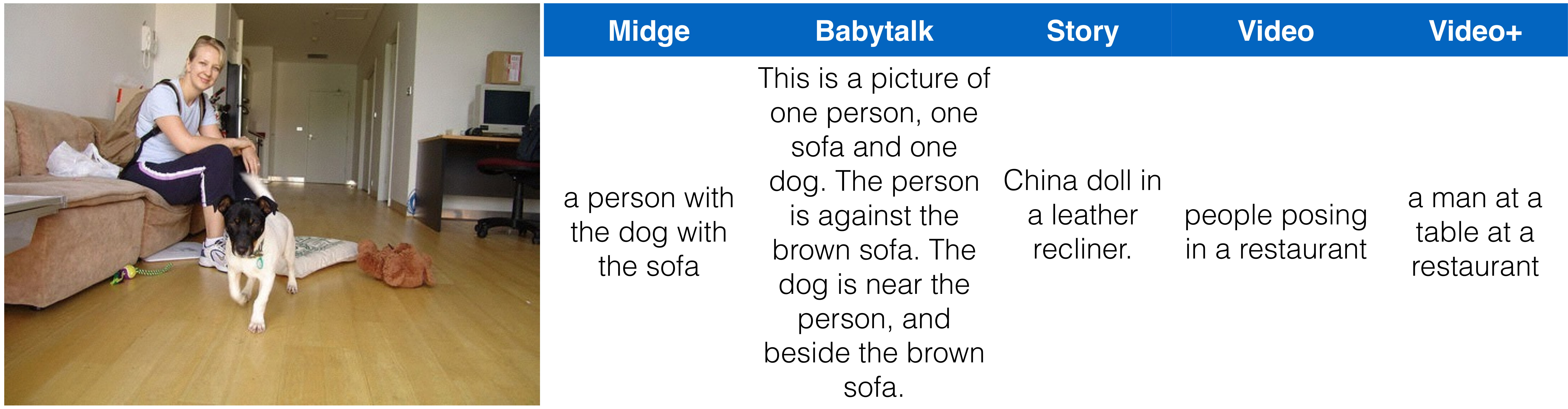}
\caption{Descriptions produced by Midge~\cite{midge}, Babytalk~\cite{babytalk}, Story~\cite{Story}, Video~\cite{rohrbach13iccv} and Video+~\cite{rohrbach13iccv} for an image. Note that since Story is a retrieval based approach, we consider the top-ranked output to show here.}
\label{fig:6}
\end{figure*}

In the paper, we compared the relative performance of five image description methods: Midge~\cite{midge}, Babytalk~\cite{babytalk}, Story~\cite{Story}, and two versions of Translating Video Content to Natural Language Descriptions~\cite{rohrbach13iccv} (Video and Video+). Here, we show a sample image with the descriptions generated by the five methods compared in the paper (Fig.~\ref{fig:6}). We can see that Midge~\cite{midge} and Babytalk~\cite{babytalk} produce the better descriptions on this image, consistent with our finding in the paper.  

\subsection*{Appendix VII : Other Metrics}
\label{sec:6}
Our goal is to automatically evaluate for an image $I_i$ how well a candidate sentence $c_i$ matches the consensus of a set of image descriptions $\RS_{i} = \{\rs_{i1},\ldots,\rs_{im}\}$. The sentences are represented using sets of \ngrams, where an \ngram~$\omega_k \in \Omega$ is a set of one or more ordered words. In this paper we explore \ngrams~with one to four words. Each word in an \ngram is modified to its stemming or root form. That is, ``fishes'', ``fishing '' and ``fished'' all get reduced to ``fish''. The number of times an \ngram~$\omega_k$ occurs in a sentence $\rs_{ij}$ is denoted $h_k(\rs_{ij})$ or $h_k(c_i)$ for the candidate sentence $c_i \in C$.
\subsubsection*{\bleu}
\label{ss:bleu}
\bleu~\cite{papineni} is a popular machine translation metric that analyzes the
co-occurrences of \ngrams~between the candidate and reference sentences. As we explain in Sec.\ref{sec:experiments}, we compute the sentence level \bleu scores between a candidate sentence and a set of reference sentences. The \bleu score is computed as follows:
\begin{equation}
	P_n(c_i , \RS_i) = \frac{\sum_{k} \min ( h_k(c_i), \max\limits_{j \in m}h_k(\rs_{ij}))}
	{\sum_{k} h_k(c_i)},
\end{equation}
where $k$ indexes the set of possible \ngrams\ of length $n$. The clipped
precision metric limits the number of times an \ngram~may be counted to the
maximum number of times it is observed in a single reference sentence. Note that
$P_n$ is a precision score and it favors short sentences. So a brevity penalty
is also used:
\begin{equation}
b(C, \RS)= \begin{cases} 1 &\text{if } l_C > l_\RS \\
	      e^{1-l_\RS/l_C} &\text{if } l_C \le l_\RS
          \end{cases},
\end{equation}
where $l_C$ is the total length of candidate sentences $c_i$'s and $l_\RS$ is
the length of the corpus-level effective reference length.
When there are multiple references for a candidate sentence, we choose to use
the {\it closest} reference length for the brevity penalty.

The overall \bleu~score is computed using a weighted geometric mean of the
individual \ngram~precision:
\begin{align}
	\bleu_{N}(c_i, \RS_i) = b(c_i, \RS_i) \exp\left( \sum_{n=1}^N w_n \log P_n(c_i, \RS_i)\right),
\end{align}
where $N = 1, 2, 3, 4$ and $w_n$ is typically held constant for all $n$.

\bleu~has shown good performance for corpus-level comparisons over which a high
number of \ngram~matches exist. However, at a sentence-level the \ngram~matches
for higher $n$ rarely occur. As a result, \bleu~performs poorly when comparing
individual sentences.
\subsubsection*{ROUGE}
\label{ss:rouge}
ROUGE is a set of evaluation metrics designed to evaluate text summarization algorithms.
\begin{enumerate}
\item \rougen:
The first \rouge~metric computes a simple \ngram~recall over all reference summaries given a candidate sentence:
\begin{equation}
ROUGE_N( c_i,\RS_{i}) = \frac{\sum_j \sum_k \min(h_k(c_i), h_k(\rs_{ij}))} {\sum_j \sum_k h_k(\rs_{ij})}
\end{equation}
\begin{figure*}[t]
\centering
\begin{subfigure}{0.33\textwidth}
\includegraphics[width=\columnwidth, page=1]{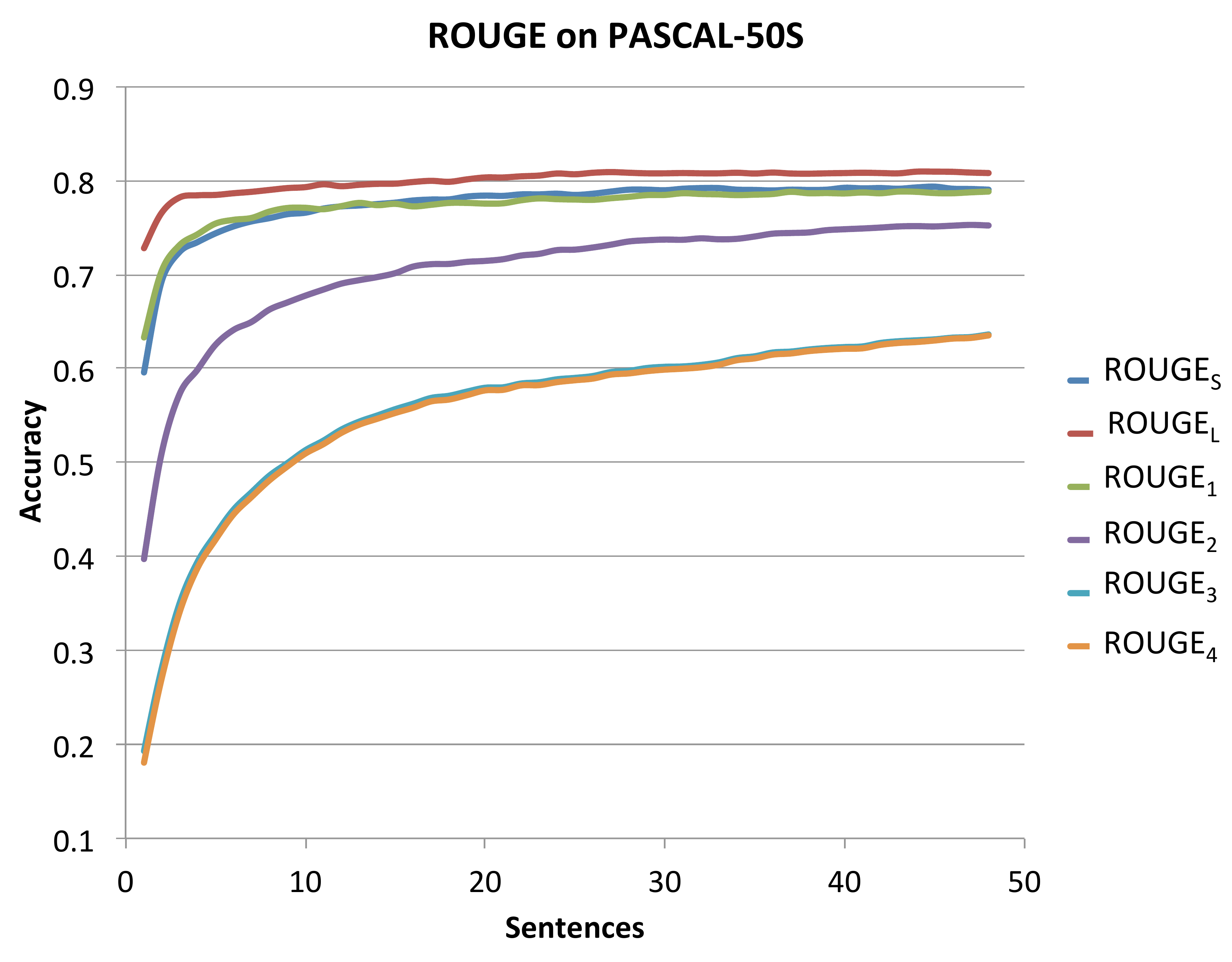}
\end{subfigure}
\begin{subfigure}{0.33\textwidth}
\includegraphics[width=\columnwidth, page=2]{figures/pascal}
\end{subfigure}
\begin{subfigure}{0.33\textwidth}
\includegraphics[width=\columnwidth, page=3]{figures/pascal}
\end{subfigure}
\caption{Performance of different versions of metrics on \pascal}
\label{fig:4}
\end{figure*}
\begin{figure*}[t]
\centering
\begin{subfigure}{0.33\textwidth}
\includegraphics[width=\columnwidth, page=4]{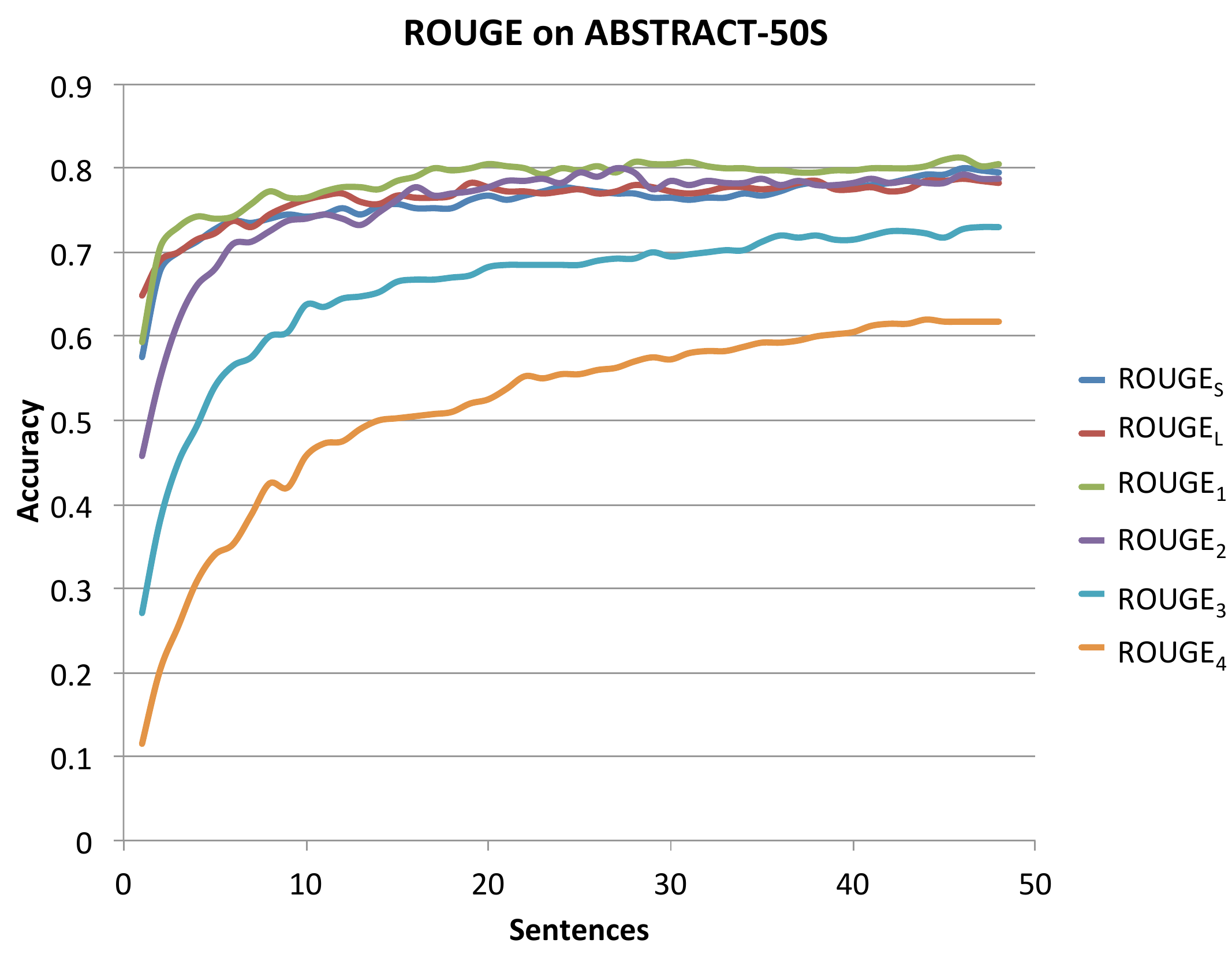}
\end{subfigure}
\begin{subfigure}{0.33\textwidth}
\includegraphics[width=\columnwidth, page=5]{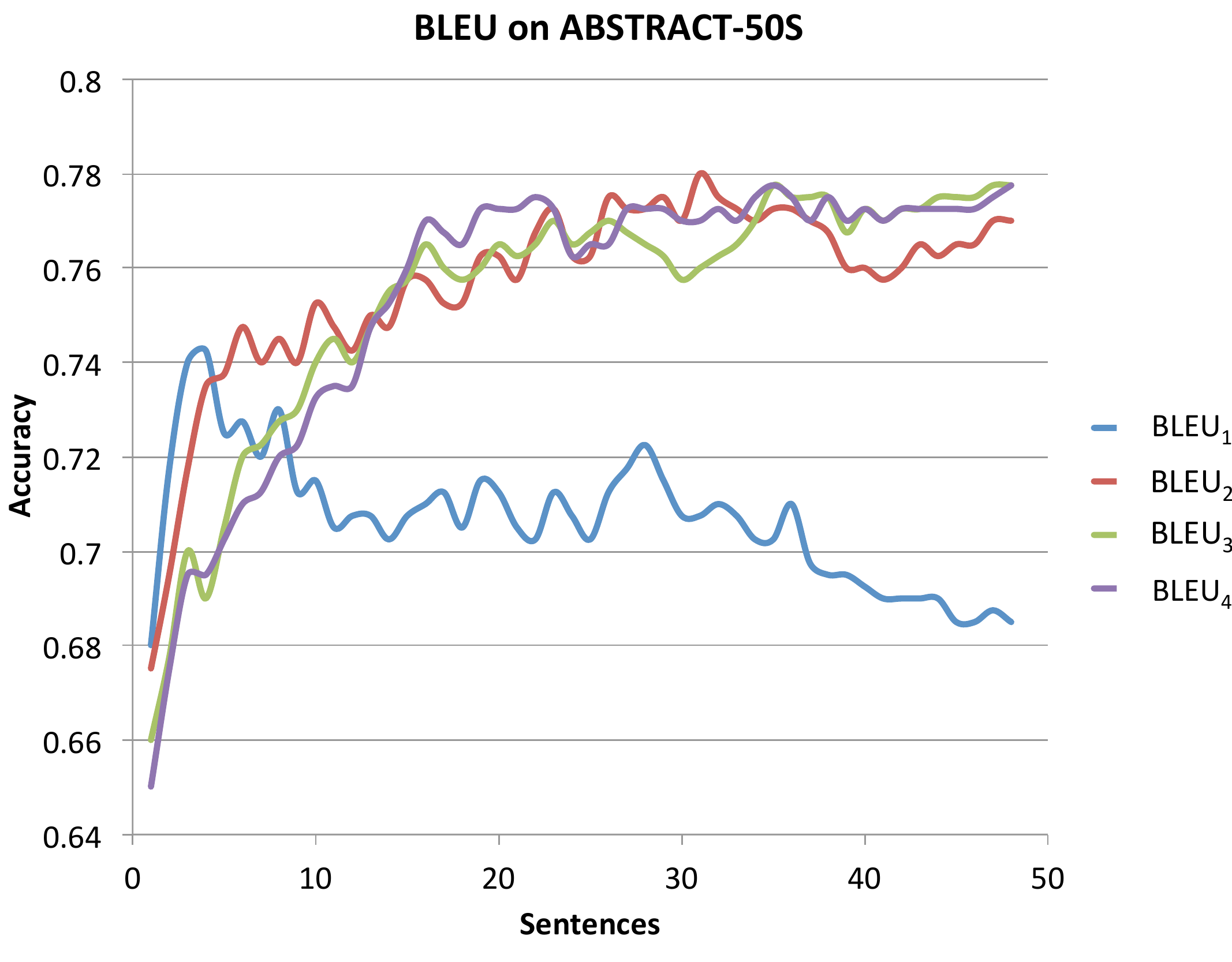}
\end{subfigure}
\begin{subfigure}{0.33\textwidth}
\includegraphics[width=\columnwidth, page=6]{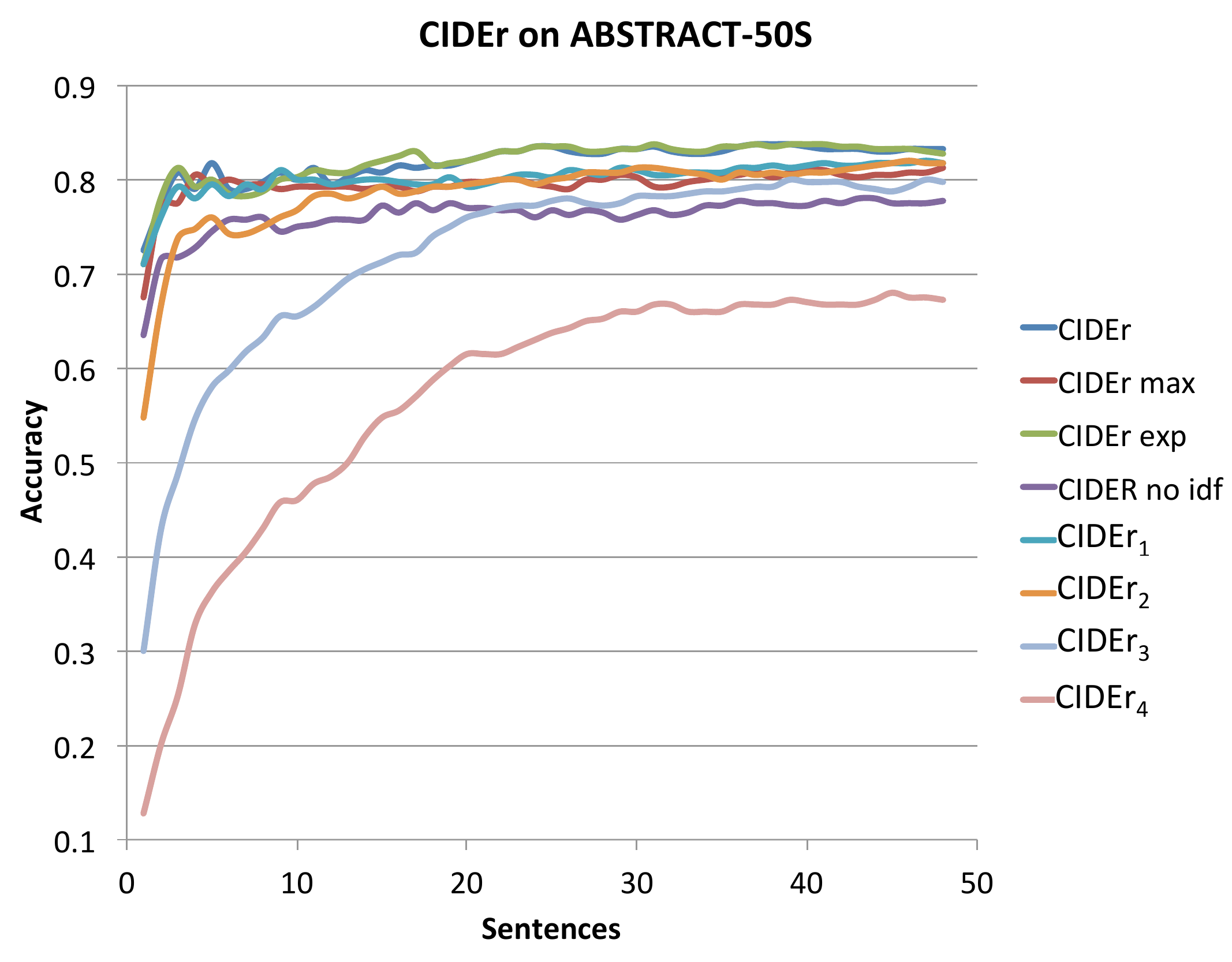}
\end{subfigure}
\caption{Performance of different versions of metrics on \abstractscenes}
\label{fig:5}
\end{figure*}

\item \rougel:
\rougel~ uses a measure based on the Longest Common Subsequence (LCS). An LCS is a set words shared by two sentences which occur in the same order. However, unlike \ngrams~there may be words in between the words that create the LCS. Given the length $l(c_i,\rs_{ij})$ of the LCS between a pair of sentences, \rougel~is found by computing an F-measure:

\begin{align} R_{l} &= \max\limits_{j}\frac{l(c_i,\rs_{ij})}{|\rs_{ij}|} \\
		P_{l} &= \max\limits_{j}\frac{l(c_i,\rs_{ij})}{|c_i|} \\
		ROUGE_L( c_i,\RS_{i}) &= \frac{(1 + \beta^2) R_{l} P_{l}}{R_{l} + \beta^2 P_{l}}
\end{align}
$R_l$ and $P_l$ are recall and precision of LCS. $\beta$ is usually set to favor \emph{recall} ($\beta = 2$). Since \ngrams~are implicit in this measure due to the use of the LCS, they need not be specified.

\item \rouges:
The final \rouge~metric uses skip bi-grams instead of the LCS or \ngrams. Skip bi-grams are pairs of ordered words in a sentence. However, similar to the LCS, words may be skipped between pairs of words. Thus, a sentence with 4 words would have $C^{4}_{2} = 6$ skip bi-grams. Precision and recall are again incorporated to compute an F-measure score. If $f_k(s_{ij})$ is the skip bi-gram count for sentence $s_{ij}$, \rouges~is computed as:
\begin{align}
R_{s} &= \max\limits_{j} \frac{\sum_k \min(f_k(c_i), f_k(\rs_{ij}))}{\sum_k f_k(\rs_{ij})} \\
P_{s} &= \max\limits_{j} \frac{\sum_k \min (f_k(c_i), f_k(\rs_{ij}))}{\sum_k f_k(c_{i})} \\
ROUGE_S( c_i,\RS_{i}) &= \frac{(1 + \beta^2) R_{s} P_{s}}{R_{s} + \beta^2 P_{s}}
\end{align}
Skip bi-grams are capable of capturing long range sentence structure. In practice, skip bi-grams are computed so that the component words occur at a distance of at most 4 from each other.
\end{enumerate}
\subsubsection*{\meteor}
\meteor is calculated by generating an alignment between the words in the candidate and reference sentences, with an aim of 1:1 correspondence. This alignment is computed while minimizing the number of chunks, $ch$, of contiguous and identically ordered tokens in the sentence pair. The alignment is based on exact token matching, followed by WordNet synonyms and then stemmed tokens. Given a set of alignments, $m$, the \meteor score is the harmonic mean of precision and recall between the best scoring reference and candidate:
\begin{align}
Pen = \gamma \left( \frac{ch}{m} \right)^\theta \\
F_{mean} = \frac{P_m R_m}{\alpha P_m + (1-\alpha) R_m} \\
P_m = \frac{|m|}{\sum_{k} h_k(c_i)}\\
R_m = \frac{|m|}{\sum_{k} h_k(s_{ij})}\\
METEOR = (1-Pen) F_{mean}
\end{align}
Thus, the final \meteor score includes a penalty based on chunkiness of resolved matches and a harmonic mean term that gives the quality of the resolved matches. 
\subsection*{Appendix VIII : Detailed Evaluation}
\label{sec:7}
We now show the results for different versions of each metric in the family of \bleu and \rouge metrics, along with some variations of \cider. We use only one (latest) version of \meteor, thus it is not a part of this evaluation. 
The versions of \cider shown here are as follows. \textbf{\cider exp} refers to an exponential combination of scores obtained by varying \ngram counts $w_n$ instead of taking a mean, which we describe in Sec.~\ref{sec:metric}. \textbf{\cider max} refers to taking a max across scores with different reference sentences, instead of the mean we discuss in the paper. \textbf{\cider no idf} version sets uniform IDF weights in \cider. The rest of the versions of other metrics are explained in the previous section. The results on \pascal are shown in Fig.~\ref{fig:4} and \abstractscenes are shown in Fig.~\ref{fig:5}. We find that removing the IDF weights in the \textbf{\cider no idf} version hurts performance significantly. \textbf{\cider max} and \textbf{\cider exp} perform slightly worse than \cider. The best performing version of each of these metrics was discussed in Sec.~\ref{sec:results}.

\end{document}